# Analysis of Watson's Strategies for Playing Jeopardy!

Gerald Tesauro                               GTESAURO@US.IBM.COM
**David C. Gondek**                             DGONDEK@US.IBM.COM
**Jonathan Lenchner**                         LENCHNER@US.IBM.COM
**James Fan**                                       FANJ@US.IBM.COM
**John M. Prager**                                JPRAGER@US.IBM.COM
*IBM TJ Watson Research Center, Yorktown Heights, NY 10598 USA*

## Abstract

Major advances in Question Answering technology were needed for IBM Watson[1] to play *Jeopardy![2]* at championship level – the show requires rapid-fire answers to challenging natural language questions, broad general knowledge, high precision, and accurate confidence estimates. In addition, *Jeopardy!* features four types of decision making carrying great strategic importance: (1) Daily Double wagering; (2) Final Jeopardy wagering; (3) selecting the next square when in control of the board; (4) deciding whether to attempt to answer, i.e., "buzz in." Using sophisticated strategies for these decisions, that properly account for the game state and future event probabilities, can significantly boost a player's overall chances to win, when compared with simple "rule of thumb" strategies.

This article presents our approach to developing Watson's game-playing strategies, comprising development of a faithful simulation model, and then using learning and Monte-Carlo methods within the simulator to optimize Watson's strategic decision-making. After giving a detailed description of each of our game-strategy algorithms, we then focus in particular on validating the accuracy of the simulator's predictions, and documenting performance improvements using our methods. Quantitative performance benefits are shown with respect to both simple heuristic strategies, and actual human contestant performance in historical episodes. We further extend our analysis of human play to derive a number of valuable and counterintuitive examples illustrating how human contestants may improve their performance on the show.

## 1. Introduction

*Jeopardy!* is a fast-paced, demanding, and highly popular TV quiz show that originated in the US, and now airs in dozens of international markets. The show features challenging questions (called "clues" in the show's parlance) drawn from a very broad array of topics; the clues may embody all manner of complex and ambiguous language, including vague allusions and hints, irony, humor and wordplay.

The rules of game play in regular episodes (Jeopardy! Gameplay, 2013) are as follows. There are two main rounds of play, wherein each round uses a board containing 30 squares, organized as five squares in six different categories, with each square containing a hidden clue. Second-round clues have higher dollar values, presumably reflecting greater difficulty. In typical play, a square will be selected according to category and dollar amount by the player in control of the board, and its clue will be revealed and read aloud by the host.

---

1. Registered trademark of IBM Corp.
2. Registered trademark of Jeopardy Productions Inc.

When the host finishes reading the clue, players may attempt to answer, i.e., "buzz in," by pressing a signaling device. The first player to buzz in obtains the right to attempt to answer; if the answer is correct, the player's score increases by the clue's dollar value, whereas if the answer is incorrect, the player's score decreases by the dollar value, and the clue is re-opened for the other players to attempt to answer on the "rebound."

One clue in the first round, and two clues in the second round have a special "Daily Double" status. The player who selected the clue has the exclusive right to answer, and she must specify a wager between $5 and either her current score or the round limit, whichever is greater. The game concludes with a single clue in the "Final Jeopardy" round. The players write down sealed-bid wagers, and then have 30 seconds to write an answer after the clue is revealed. The player finishing with the highest dollar-value score[3] wins that amount, and can continue playing on the next episode.

When IBM began contemplating building a computing system to appear on *Jeopardy!* it was readily apparent that such an undertaking would be a hugely daunting challenge for automated Question Answering (QA) technology. State-of-the-art systems at that time were extremely poor at general open-domain QA, and had considerable difficulty if the questions or the supporting evidence passages were not worded in a straightforward way. Building the DeepQA architecture, and advancing its performance at *Jeopardy!* to a competitive level with human contestants, would ultimately require intense work over a four-year period by a team of two dozen IBM Researchers (Ferrucci, Brown, Chu-Carroll, Fan, Gondek, Kalyanpur, ..., & Welty, 2010; Ferrucci, 2012).

Rather than discussing WATSON's QA performance, which is amply documented elsewhere, the purpose of this paper is to address an orthogonal and significant aspect of winning at *Jeopardy!*, namely, the strategic decision-making required in game play. There are four types of strategy decisions: (1) wagering on a Daily Double (DD); (2) wagering in Final Jeopardy (FJ); (3) selecting the next square when in control of the board; and (4) deciding whether to attempt to answer, i.e., "buzz in." The most critical junctures of a game often occur in the Final Jeopardy round and in playing Daily Doubles, where wagering is required. Selecting a judicious amount to wager, based on one's confidence, the specific game situation, and the likely outcomes of the remaining clues, can make a big difference in a player's overall chance to win. Also, given the importance of Daily Doubles, it follows that a player's square selection strategy when in control of the board should result in a high likelihood of finding a DD. Allowing one's opponents to find the DDs can lead to devastating consequences, especially when playing against Grand Champions of the caliber of Ken Jennings and Brad Rutter. Furthermore, a contestant's optimal buzz-in strategy can change dramatically in certain specific end-game scenarios. For example, a player whose score is just below half the leader's score may need to make a "desperation buzz" on the last clue in order to avoid a guaranteed loss. Conversely, at just above half the leader's score, the correct strategy may be to never buzz in.

There is scant prior literature on *Jeopardy!* game strategies, and nearly all of it is qualitative and heuristic, with the sole exception of Final Jeopardy strategy, where substantial quantitative analysis is embodied in the J! Archive (2013) Wagering calculator. Additionally, Dupee (1998) provides a detailed analysis of betting in Final Jeopardy, with particular emphasis on the so-called "two-thirds" scenario, where betting nothing may increase winning chances for the second-place player, provided that he has at least two-thirds of the leader's score. Some qualitative guidelines for aggressive or conservative Daily Double betting are also given, depending on confidence in the category, ability to win the buzz, and positioning for Final Jeopardy. A specific last-clue DD bet is presented where the best bet either takes the lead, or drops to exactly half the leader's score (i.e., a "lock-tie"), resulting in extra chances to win. Harris (2006), one of the show's top contestants, provides numerous qualitative insights into strategic thinking at championship level, including the importance of seeking DDs in the bottom rows, wagering to position for Final Jeopardy, and protecting a lead late in the game by being cautious on the buzzer.

---

3. Multiple players may win if they finished tied for first place. This deviation from a strict zero-sum game can lead to fascinating counter-intuitive strategies.

This article describes our team's work in developing a collection of game-strategy algorithms deployed in Watson's live *Jeopardy!* contests against human contestants. To our knowledge, this work constitutes the first-ever quantitative and comprehensive approach to *Jeopardy!* strategy that is explicitly based on estimating and optimizing a player's probability of winning in any given *Jeopardy!* game state. Our methods enable Watson to find DDs faster than humans, and to calculate optimal wagers and buzz-in thresholds to a degree of precision going well beyond human capabilities in live game play. A brief overview of our work (Tesauro, Gondek, Lenchner, Fan, & Prager, 2012) recently appeared in a special issue of the IBM Journal of Research and Development. The present article provides expanded descriptions of each of our strategy algorithms, and presents substantial new quantitative documentation of the performance advantages obtained by our approach, when compared both to simple heuristic strategies as well as to actual human strategies.

The overall organization of the article is as follows. We first provide in section 1.1 a glossary of important technical terms and notation used throughout the article. Section 2 then overviews our general approach to developing a *Jeopardy!* simulator, which we use to simulate contests between Watson and human contestants. Studying game-playing programs in simulation is a well-established practice in computer games research. However, modeling *Jeopardy!* is a much more difficult undertaking than in traditional games like Checkers and Chess, due to its rich language content and extensive imperfect information. It is essential to model the statistical performance profiles of human contestants, as well as their tendencies in wagering and square selection, by mining historical data on contestant performance in thousands of previously aired episodes. In this respect, *Jeopardy!* is similar to other imperfect-information games like Poker (Billings, Davidson, Schaeffer, & Szafron, 2002), where effective dynamic modeling of one's opponents is a requisite ingredient for strong play by both computers and humans. The general overview is followed in sections 2.1-2.6 by specific designs and construction methodologies for our simulation component models, emulating Daily Double placement, human performance in Daily Doubles, Final Jeopardy, regular clues and square selection, as well as extensions of such models from single-game to multi-game format. The modeling section concludes in section 2.7, which presents a statistically meaningful validation study, documenting how well various game statistics predicted by the simulator match up with actual statistics in live matches between Watson and human contestants. As detailed in Appendix 1, Watson played more than 100 such "Sparring Games" before appearing on television, and the validation study specifically focuses on the final 55 games in which Watson faced off against extremely strong former *Jeopardy!* champions.

Section 3 presents specific techniques for designing, learning and optimizing WATSON's four strategy modules over the course of many simulated games. These techniques span a range of widely used methods in current AI/OR research studies. Specifically, section 3.1 details our approach to DD wagering, which combines nonlinear regression with Reinforcement Learning to train a Game State Evaluator over the course of millions of simulated games. Section 3.2 presents methods to calculate a Best-Response wagering strategy (a standard game-theoretic concept) in Final Jeopardy using either offline or online Monte-Carlo sampling. Section 3.3 describes WATSON's square selection strategy, the most important ingredient of which is live Bayesian inference calculation of the probabilities of various squares containing a Daily Double. Finally, section 3.4 documents how WATSON computes buzz-in thresholds in endgame states using a combination of Approximate Dynamic Programming with online Monte-Carlo trials, i.e., "rollouts" (Tesauro & Galperin, 1996; Bertsekas & Castanon, 1999).

As our work has led to many new insights into what constitutes effective *Jeopardy!* strategy, section 4 of the paper presents some of the more interesting and counterintuitive insights we have obtained, with the hope of improving human contestant performance. Section 4.1 gives an overview of the most important decision boundaries we found in WATSON's Best-Response FJ strategy. Section 4.2 discusses our most important finding regarding human DD wagering, namely that humans should generally bet more aggressively. Section 4.3 presents buzz threshold analysis yielding initial buzz thresholds that are surprisingly aggressive, and rebound thresholds that are surprisingly conservative. Finally, section 4.4 discusses unusual and seemingly paradoxical implications of the "lock-tie" FJ scenario, where the leader must bet $0 to guarantee a win.

After summarizing our work and lessons learned in section 5, Appendix 1 provides details on WATSON's competitive record, and Appendix 2 gives mathematical details of the buzz threshold calculation.

## 1.1 Glossary

In this section we provide definitions of various technical terms and notation used in subsequent sections to describe our simulation models, strategies, or aspects of *Jeopardy!* game play.

- A, B and C - The players with the highest, second highest and third highest scores, respectively, or their current scores.

- Accuracy - The probability that a player will answer a clue correctly, in situations where answering is mandatory (Daily Doubles or Final Jeopardy).

- Anti-two-thirds bet - Potential counter-strategy for A in the Two-thirds Final Jeopardy scenario (see "Two-thirds bet"). After a two-thirds bet, B's score is at most 4B-2A. This will be less than A if B is less than three-fourths of A. Hence, A could guarantee a win by small bet of at most 3A-4B. However, such a bet is vulnerable to B making a large bet that significantly overtakes A.

- Average Contestant model - A model based on aggregate statistics of all J! Archive regular episode data (excluding special-case contestant populations).

- Bet to cover, Shut-out bet - Standard strategy for the leader, A, in Final Jeopardy. A usually bets at least 2B-A (frequently 2B-A+1). After a correct answer, A's score is at least 2B, which guarantees a win.

- Board Control - The right to select the next clue; usually belonging to the player who gave the last correct answer. Daily Doubles can only be played by the player with control of the board.

- Buzz attempt rate - Parameter $b$ in the regular clue simulation model, denoting the average probability that a player will attempt to buzz in on a clue.

- Buzz correlation - In the regular clue model, a quantity $\rho_{ij}$ indicating the degree to which the buzz-in decisions of player $i$ and $j$ tend to be the same on a given clue (see "Correlation coefficient"). For two humans, $\rho_{ij} = \rho_b$ (empirically $\sim 0.2$), whereas $\rho_{ij} = 0$ between a human and WATSON.

- Buzzability - Short for "buzzer ability." The probability of a given player winning a contested buzz when multiple players buzz in.

- Buzzing In - Pressing a signaling device, indicating that a player wishes to attempt to answer a clue. After the host finishes reading the clue, the first player to buzz in is allowed to answer.

- Champion model - A model based on aggregate statistics of the 100 best players in the J! Archive dataset, ranked according to number of games won.

- Correlation coefficient - In the simulator, a quantity $\rho_{ij}$ indicating the degree to which randomized binary events (Buzz/NoBuzz or Right/Wrong) for players $i$ and $j$ tend to be the same on a given clue. As a simple example, suppose $i$ and $j$ have 50% chance each of answering Final Jeopardy correctly. Let $P(x_i, x_j)$ denote the joint probability that $i$ has correctness $x_i$ and $j$ has correctness $x_j$. The correlation coefficient is then given by $\rho_{ij} = P(\text{Right,Right}) + P(\text{Wrong,Wrong}) - P(\text{Right,Wrong}) - P(\text{Wrong,Right})$. Note that if $x_i$ and $x_j$ are independent, then all four joint outcomes are equally likely, so that $\rho_{ij} = 0$. If $x_i$ and $x_j$ always match, then $\rho_{ij} = 1$ and if they always mismatch, then $\rho_{ij} = -1$.

- Exhibition Match - (See Appendix 1) The televised two-game match, aired in Feb. 2011, in which WATSON competed against Brad Rutter and Ken Jennings, arguably the two best-ever human contestants (see "Multi-game format").

- Grand Champion model - A model based on aggregate statistics of the ten best players in the J! Archive dataset, ranked according to number of games won.

- Lockout, locked game - A game state in which the leader's current score cannot be surpassed by the opponents in play of the remaining clues, so that the leader has a guaranteed win. Usually refers to Final Jeopardy states where the leader has more than double the opponents' scores.

- Lock-tie - A Final Jeopardy situation in which the player in second place has exactly half the leader's score. The leader has a guaranteed win by betting $0, enabling the second place player to achieve a tie for first place by betting everything and answering correctly.

- Match equity - Objective function optimized by Watson in the two-game Exhibition Match, defined as probability of finishing first plus 0.5 times probability of finishing second. By contrast, Watson simply maximized probability of winning in the Sparring Games.

- Multi-game format - A special-case format used in the finals of the Tournament of Champions, and in the Exhibition Match with Ken Jennings and Brad Rutter. First, second and third place are awarded based on point totals over two games. In the event of a first-place tie, a sudden death tie-break clue is played. Depending on the prize money, there can be a significant incentive to finish in second place.

- QA - Short for Question Answering. A computing system or a suite of Natural Language Processing techniques used to search for, evaluate, and select candidate answers to clues.

- Precision, Precision@$b$ - For regular clues, the average probability that a player will answer correctly on the fraction of clues ($b$) in which the player chooses to buzz in and answer (Ferrucci et al., 2010).

- Rebound - The situation after the first player to buzz in gets the clue wrong, and the remaining players then have another chance to buzz in and try to answer.

- Regular episode format - In regular episodes, a returning champion plays a single game against two new challengers. First, second and third place are determined by point totals in that game, and multiple players may finish tied for first. The player(s) finishing first will continue to play in the next episode. There is little incentive to finish second, as it only pays $1000 more than finishing third.

- Right/wrong correlation - In the regular clue model, a quantity $\rho_{ij}$ indicating the degree to which the correctness of player $i$ and $j$ tend to be the same on a given clue (see "Correlation coefficient"). For two humans, $\rho_{ij} = \rho_p$ (empirically $\sim 0.2$), whereas $\rho_{ij} = 0$ between a human and Watson.

- Rollouts - Extensive Monte-Carlo simulations used to estimate the probability of a player winning from a given game state.

- Sparring Games - (See Appendix 1) Two series of practice games (Series 1, 2) played by Watson against former *Jeopardy!* contestants. Series 1 games were against contestants selected to be typical of "average" contestants appearing on the show. Series 2 games were played against "champions," i.e., contestants who had reached the finals or semi-finals of the annual Tournament of Champions.

- Tip-off effect - Information revealed in an initial incorrect answer that helps the rebound player deduce the correct answer. For example, a clue asking about a "New

Jersey university" is likely to have only two plausible answers, Rutgers and Princeton. After an initial answer "What is Princeton?" is ruled incorrect, the rebounder can be highly confident that the correct answer is "What is Rutgers?"

- Two-thirds bet - A plausible Final Jeopardy strategy for B in cases where B has at least two-thirds of A's score. Assuming A makes a standard bet to cover of at least 2B-A, B's winning chances are optimized by betting at most 3B-2A. With such a bet, B will win whenever A is wrong, whereas for larger bets, B also needs to be right. This strategy is vulnerable to a counter-strategy by A (see "Anti-two-thirds bet").

## 2. Simulation Model Approach

Since we optimize Watson's strategies over millions of synthetic matches, it is important that the simulations be faithful enough to give reasonably accurate predictions of various salient statistics of live matches. Developing such a simulator required significant effort, particularly in the development of human opponent models.

The use of a simulator to optimize strategies is a well-established practice in computer games research. Simulated play can provide orders of magnitude more data than live game play, and does not suffer from overfitting issues that are commonly encountered in tuning to a fixed suite of test positions. While it is usually easy to devise a perfect model of the rules of play, simulation-based approaches can face a significant challenge if accurate models of opponent strategies are required. In traditional two-player zero-sum perfect-information board games (Backgammon, Checkers, Chess, etc.) such modeling is normally not required – one can simply aim to compute the minimax-optimal line of play, as there is limited potential to model and exploit suboptimal play by the opponents. By contrast, in repeated normal-form games such as Prisoner's Dilemma and Rock-Paper-Scissors, a one-shot equilibrium strategy is trivial to compute but insufficient to win in tournament competitions (Axelrod, 1984; Billings, 2000). The best programs in these games employ some degree of adaptivity and/or modeling based on the observed behaviors of their opponents. Poker is another prominent game where opponent modeling is essential to achieve strong play (Billings et al., 2002) . Playing an equilibrium strategy when the opponent is bluffing too much or too little would forego an opportunity to significantly boost one's expected profit.

In contrast to the above-mentioned games, the *Jeopardy!* domain introduces entirely new modeling issues, arising from the natural language content in its category titles, clues, and correct answers. Obviously our simulator cannot generate synthetic clues comparable to those written by the show's writers, nor can we plausibly emulate actual contestant responses. Even the most basic analysis of categories and clues (e.g., which categories tend to be "similar," co-occurrence likelihood of categories in a board, what type of information is provided in clues, what type of mental process is needed to infer the correct response, how the clue difficulty is calibrated based on the round and dollar value) seemed daunting and the prospects for success seemed remote. Likewise, modeling the distributions of human contestant capabilities over thousands of categories, and correlations of abilities across different categories, seemed equally implausible.

Due to the above considerations, our initial simulator design was based on an extremely simplified approach. We avoided any attempt to model the game's language content, and decided instead to devise the simplest possible stochastic process models of the various events that can occur at each step of the game. Our plan was to examine how accurately such a simulator could predict the outcomes of real Watson-vs-human *Jeopardy!* matches, and refine the models as needed to correct gross prediction errors. As it turned out, our simple simulation approach predicted real outcomes much more accurately than we initially anticipated (see section 2.7 below), so that no major refinements were necessary.

The only noteworthy enhancement of our simple stochastic process models occurred in 2010, after Watson had acquired the ability to dynamically learn from revealed answers in a category (Prager, Brown, & Chu-Carroll, 2012). The effect was substantial, as Watson's accuracy improved by about 4% from the first clue to the last clue in a category. We captured this effect by using historical data: each category in a simulated game would be paired with a randomly drawn historical category, where a sequence of five right/wrong Watson answers was known from prior processing. Instead of stochastically generating right/wrong answers for Watson, the simulator used the recorded sequence, which embodied the tendency to do better on later clues in the category. The ability to simulate this learning effect was instrumental in the ultimate development of Watson's square selection algorithm, as we describe in section 3.3.

Our stochastic process simulation models are informed by:

- (i) properties of the game environment (rules of play, DD placement probabilities, etc.)

- (ii) performance profiles of human contestants, including tendencies in wagering and square selection;

- (iii) performance profiles of Watson, along with Watson's actual strategy algorithms;

- (iv) estimates of relative "buzzability" of Watson vs. humans, i.e., how often a player is able to win the buzz when two or more contestants are attempting to buzz in.

Our primary source of information regarding (i) and (ii) is a collection of comprehensive historical game data available on the J! Archive (2013) web site. We obtained fine-grained event data from approximately 3000 past episodes, going back to the mid-1990s, annotating the order in which clues were played, right/wrong contestant answers, DD and FJ wagers, and the DD locations. After eliminating games with special-case contestants (Teen, College, Celebrity, etc. games), the remaining data provided the basis for our model of DD placement (section 2.1), and models of human contestant performance in Daily Doubles (section 2.2), Final Jeopardy (section 2.3), regular clues (section 2.4), and square selection (section 2.5).

We devised three different versions of each human model, corresponding to three different levels of contestant ability encountered during Watson's matches with human contestants (see Appendix 1 for details). The "Average Contestant" model was fitted to all non-tournament game data – this was an appropriate model of Watson's opponents in the Series 1 sparring games. The "Champion" model was designed to represent much stronger opponents that Watson faced in the Series 2 sparring games; we developed this model using data from the 100 best players in the dataset, ranked by number of games won. Finally, for our Exhibition Match with Ken Jennings and Brad Rutter, we devised a "Grand Champion" model which was informed by performance metrics of the 10 best players. Since the Exhibition Match used a multi-game format (1st, 2nd and 3rd place determined by two-game point totals), we developed specialized DD and FJ wagering models for Game 1 and Game 2 of the match, as described in section 2.6.

## 2.1 Daily Double Placement

We calculated the joint row-column frequencies in the J! Archive data of Round 1 and Round 2 DD placement; the Round 2 frequencies are illustrated in Figure 1. Our analysis confirms well-known observations that DDs tend to be found in the lower rows (third, fourth and fifth) of the board, and basically never appear in the top row. However, we were surprised to discover that there are also column dependencies, i.e., some columns are more likely to contain a DD than others. For example, DDs are most likely to appear in the first column, and least likely to appear in the second column. (We can only speculate[4] why the show's producers place DDs in this fashion.)

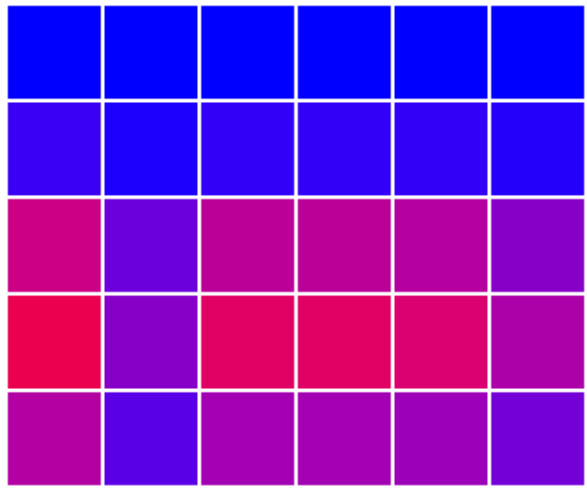

Figure 1: Illustration of row-column frequencies of second-round DD placement in ∼ 3000 previous *Jeopardy!* episodes. Red denotes high frequency and blue denotes low frequency.

Additional analytic insights from the data include: (i) The two second-round DDs never appear in the same column. (ii) The row location appears to be set independently of the column location, and independently of the rows of other DDs within a game. (iii) The Round 2 column-pair statistics are mostly consistent with independent placement, apart from the constraint in (i). However, there are a few specific column pair frequencies that exhibit borderline statistically significant differences from an independent placement model.

Based on the above analysis, the simulator assigns the DD location in Round 1, and the first DD location in Round 2, according to the respective row-column frequencies. The remaining Round 2 DD is assigned a row unconditionally, but its column is assigned conditioned on the first DD column.

---

4. We noted that the second column often features "pop-culture" categories (TV shows, pop music, etc.) which could account for its relative paucity of DDs.

## 2.2 Daily Double Accuracy/Betting Model

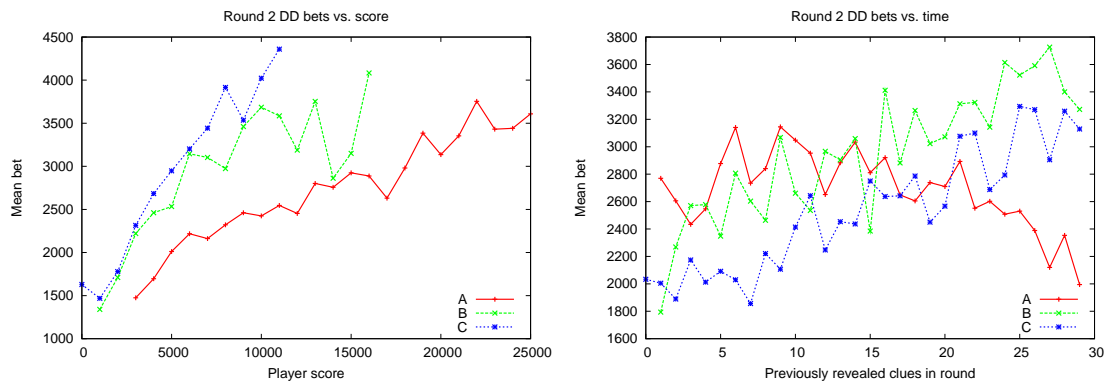

Figure 2: Average Round 2 mean DD bets of human contestants in first place (A), second place (B) and third place (C), as a function of player score (left), and clues played in round (right).

Based on the appropriate historical statistics in our J! Archive regular episode dataset, we set the mean DD accuracy parameter in our human contestant models at 64% for Average Contestants, 75% for Champions, and 80.5% for Grand Champions. Bets made by human contestants tend to be round number bets such as $1000 or $2000, and rarely exceed $5000. The main dependencies we observed are that players in the lead tend to bet more conservatively, and become extremely conservative near the end of the game, presumably to protect their lead going into Final Jeopardy. These dependencies are clearly seen in Figure 2, where we plot average bets as functions of player score and of clues played in the second round.

While the above wagering tendencies were built into our Average Contestant model, we surmised (correctly as it turned out) that much stronger Champion and Grand Champion players would quickly realize that they need to bet DDs extremely aggressively when playing against Watson. These models therefore employed an aggressive heuristic strategy which would bet nearly everything, unless a heuristic formula indicated that the player was close to a mathematically certain win.

## 2.3 Final Jeopardy Accuracy/Betting Model

The historical dataset obtained from J! Archive reveals that mean human accuracy in answering Final Jeopardy correctly is approximately 50% for average contestants, 60% for Champions, and 66% for Grand Champions. Furthermore, from statistics on the eight possible right/wrong triples, it is also clear that accuracy is positively correlated among contestants, with a correlation coefficient $\rho_{FJ} \sim 0.3$ providing the best fit to the data. We use these parameter values in simulating stochastic FJ trials, wherein we implement draws of three correlated random binary right/wrong outcomes, with means and correlations tuned to the appropriate values. This is performed by first generating correlated real numbers using a multi-variate normal distribution, and then applying suitably chosen thresholds to convert to 0/1 outcomes at the desired mean rates (Leisch, Weingessel, & Hornik, 1998). As many such draws are required to determine the precise win rate of a given FJ score combination, we also implement a lower-variance simulation method. Rather than generating a single stochastic outcome triple, the simulator evaluates all eight outcome triples, weighted by analytically derived probabilities for each combination.

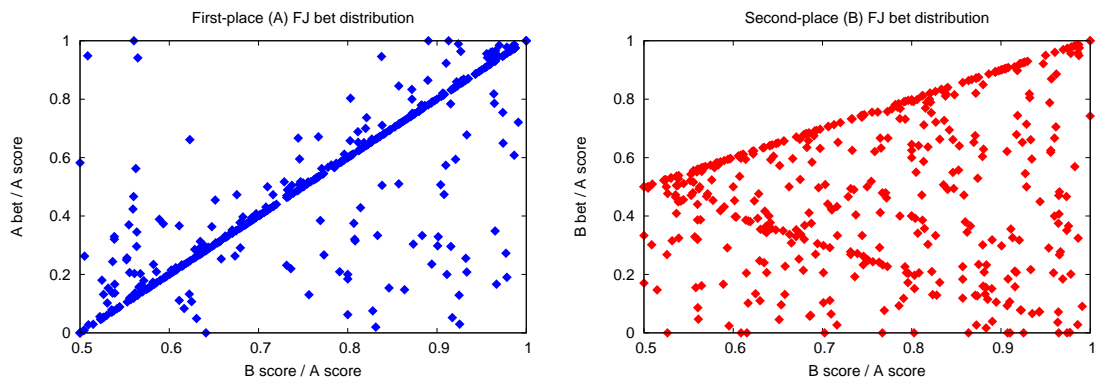

Figure 3: Distribution of human FJ bets by first-place player "A" (left) and second-place player "B" (right), normalized by leader score, as a function of B/A ratio.

The most important factor in FJ wagering is score positioning, i.e., whether a player is in first place ("A"), second place ("B") or third place ("C"). To develop stochastic process models of likely contestant bets, we first discarded data from "lockout" games (where the leader has a guaranteed win), and then examined numerous scatter-plots such as those shown in Figure 3. We see a high density line in A's bets corresponding to the well-known strategy of betting to cover in case B's score doubles to 2B. Likewise, there are two high density lines in the plot of B's bets, one where B bets everything, and one where B bets just enough to overtake A. Yet there is considerable apparent randomization apart from any known deterministic wagering principles.

After a thorough examination, we decided to segment the wagering data into six groups: we used a three-way split based on strategic breakpoints (as detailed in section 4.1) in B's score relative to A's score (less than 2/3, between 2/3 and 3/4, and more than 3/4), plus a binary split based on whether or not B has at least double C's score. We then devised wagering models for A, B, and C[5] that choose among various types of betting logic, with probabilities based on observed frequencies in the data groups. As an example, our model for B in the case (B ≥ 3/4 A, B ≥ 2C) bets as follows: bet "bankroll" (i.e., nearly everything) with 26% probability, "keepout C" (i.e., just below B-2C) with 27% probability, "overtake A" (i.e., slightly above A-B) with 15% probability, "two-thirds limit" (i.e., just below 3B-2A) with 8% probability, and various types of random bets with the remaining 24% probability mass.

---

5. Curiously enough, we saw no evidence that C's wagers vary with strategic situation, so we implemented a single betting model for C covering all six groups.

|   | Real | Model |
|---|------|-------|
| A | 65.3% | 64.8% |
| B | 28.2% | 28.1% |
| C | 7.5% | 7.4% |

Table 1: Comparison of actual human win rates with model win rates by historical replacement in 2092 non-locked FJ situations from past episodes.

The betting models described above were designed solely to match human bet distributions, and were not informed by human FJ win rates. However, we subsequently verified by a historical replacement technique that the models track actual human win rates quite closely, as shown in Table 1. We first measured the empirical win rates of the A, B, C roles[6] in 2092 non-locked FJ situations from past episodes. We then took turns recalculating the win rate of one role after replacing the bets of that role by the bet distribution of the corresponding model. The models match the target win rates very well, considering that the human bets are likely to reflect unobservable confidence estimates given the FJ category.

While we were satisfied that our human FJ model accurately fit the historical data, there was nevertheless room for doubt as to how accurately it would predict human behavior in the Sparring Games. Most notably, each of the six data groups used to estimate model parameters contained only a few hundred samples, so that the error bars associated with the estimated values were likely to be large. We could have addressed this by performing so-called second order Monte-Carlo trials (Wu & Tsang, 2004), using Gaussian draws of parameter values in each FJ trial instead of constant values, but we were concerned about significantly higher computational overhead of such an approach. There were also possibilities that contestant behavior might be non-stationary over time, which we did not attempt to model, or that contestants might alter their behavior specifically to play against WATSON. As we discuss later in section 3.2, we generally accepted the FJ model as a basis for optimizing WATSON's decisions, with the sole exception of the case where WATSON is A, and the B player may plausibly make a "two-thirds bet" (see Glossary definition in section 1.1).

## 2.4 Regular Clue Model

Our stochastic process model of regular (non-DD) clues generates a random correlated binary triple indicating which players attempt to buzz in, and a random correlated binary triple indicating whether or not the players have a correct answer. In the case of a contested buzz, a buzz winner is randomly selected based on the contestants' relative "buzzability" (ability to win a contested buzz, assumed equal in all-human matches). As mentioned in the Glossary of section 1.1, the buzz-in outcomes are governed by two tunable parameters, mean "buzz attempt rate" $b$ and "buzz correlation" $\rho_b$. The right/wrong outcomes are likewise governed by two parameters, mean "precision@$b$" (Ferrucci et al., 2010) or simply mean "precision" $p$, and "right/wrong correlation" $\rho_p$. We set the four parameter values $b$, $\rho_b$, $p$, and $\rho_p$ by running extensive Monte-Carlo simulations for many different parameter combinations, and selecting the combination yielding the best fit to observed historical frequencies of the seven possible outcomes for regular clues, as depicted in Figure 4. The outcome statistics are derived from J! Archive records of more than 150K regular clues. The parameter values we obtained for average contestants are: $b = 0.61$, $\rho_b = 0.2$, $p = 0.87$ and $\rho_p = 0.2$. The right/wrong correlation is derived directly from rebound statistics, and is particular noteworthy: while a positive value is reasonable, given the correlations seen in FJ accuracy, it might be surprising due to the "tip-off" effect on rebounds. When the first player to buzz gives a wrong answer, this eliminates a plausible candidate and could significantly help the rebound buzzer to deduce the right answer. We surmise that the data may reflect a knowledge correlation of $\sim 0.3$ combined with a tip-off effect of $\sim -0.1$ to produce a net positive correlation of 0.2.

---

6. The human win rates sum to 101%, reflecting ~1% chance of a first-place tie.

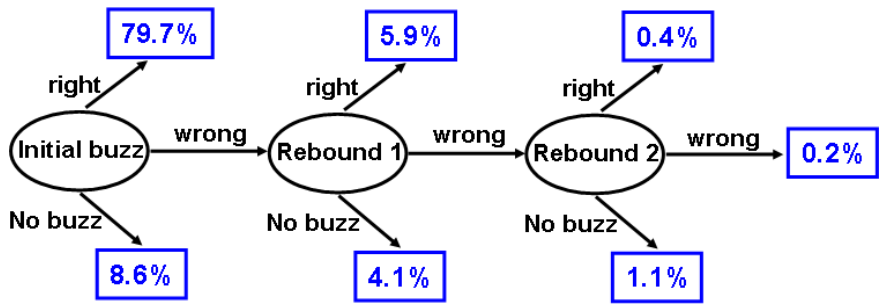

Figure 4: Frequencies of seven possible regular-clue outcomes in J! Archive average contestant dataset.

In the Champion model, there is a substantial increase in attempt rate ($b = 0.80$) and a slight increase in precision ($p = 0.89$). In the Grand Champion model, we estimated further increases in these values, to $b = 0.855$ and $p = 0.915$ respectively. As depicted in Figure 5, we also developed a refined model by segregating the regular clue data according to round and dollar value (i.e., row number), and estimating separate $(b, p)$ values in each case. Such refinements make the simulations more accurate, but do not meaningfully impact the optimization of Watson's wagering and square selection strategies. While we expect that a slight improvement in Watson's endgame buzzing could have been achieved using separate $(b, p)$ values, there was insufficient data for Champions and Grand Champions to estimate such values. Hence the deployed algorithm used constant $(b, p)$ values for all clues.

## 2.5 Square Selection Model

Most human contestants tend to select in top-to-bottom order within a given category, and they also tend to stay within a category rather than jumping across categories. There is a further weak tendency to select categories moving left-to-right across the board. Based on these observations, and on the likely impact of Watson's square selection, we developed an Average Contestant model of square selection which stays in the current category with 60% probability, and otherwise jumps to a random different category. When picking within a category there is high probability (∼90%) of picking the topmost available square. By contrast, we model Champion and Grand Champion square selection as DD seeking based on the known row statistics of DD placement. Strong players generally exhibit more Daily Double seeking when selecting squares, and when playing against Watson, they quickly adopt overt DD seeking behavior.

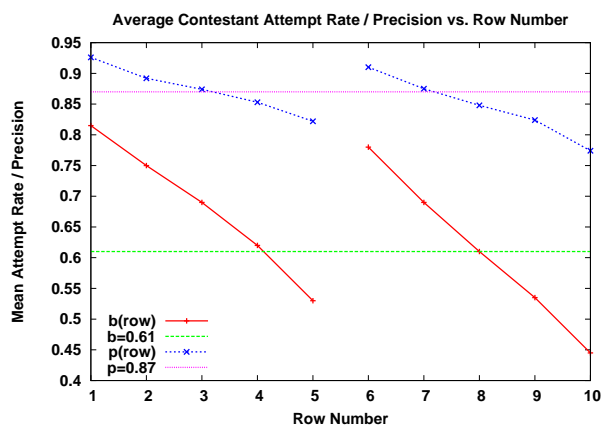

Figure 5: Estimates of Average Contestant buzz attempt rate (*b*) and precision (*p*) as a function of round and row number. Rows 1-5 denote first round clues, and rows 6-10 denote second round clues.

### 2.6 Multi-game Wagering Model

In most *Jeopardy!* contests, the winner is determined by performance in a single game. However, the show also conducts several annual tournaments, such as the Tournament of Champions, in which the final match utilizes point totals over two games to determine first, second and third place. This clearly implies that wagering strategies must differ in Game 1 and Game 2 of the match, and both need to be different from single-game wagering.

Since there is very limited multi-game match data available from J! Archive (only about two dozen Tournament of Champions final matches), it would be quite difficult to model the expected wagering of Jennings and Rutter in our Exhibition Match purely from historical data. Fortunately, we were able to make some educated guesses that considerably simplified the task. First, we predicted that they would wager DDs very aggressively in both games, unless they had an overwhelming lead. This implied that we could continue to use the aggressive heuristic DD model for single games, with a revised definition of what constitutes an "overwhelming" match lead. Second, we also expected them to bet very aggressively in Final Jeopardy of the first game. This meant that we could treat Game 1 FJ as if it were a DD situation, and again use the revised aggressive heuristic model.

The only situation requiring significant modeling effort was Game 2 FJ. We generalized the definition of "A," "B," and "C" roles for matches, based on the sum of Game 1 score plus two times Game 2 score. With this definition, all of the established single-game strategies for A, B and C carry over to two-game matches.

Given the limited available match data, only crude estimates could be assigned of the probabilities of various betting strategies. However, it is clear from the data that the wagering of human champions is much more coherent and logical than the observed wagering in regular episodes, and champion wagers frequently satisfy multiple betting constraints. These observations guided our development of revised betting models for Game 2 FJ. As an example, in the case where B has a legal generalized two-thirds bet (suitably defined for two-game matches), and B can also keep out C, our model for B bets as follows: "bankroll" bet with 35% probability, bet a small random amount that satisfies both the two-thirds and keepout-C limits with 43% probability, or bet to satisfy the larger of these two limits with 22% probability.

## 2.7 Model Validation

Our first efforts to validate the simulator's predictions occurred about half-way through Watson's first series of Sparring Games. At that point in time, the simulator had only been used to develop Watson's Final Jeopardy wagering algorithm, so the simulator was basically running a model of Watson with heuristic strategies against the Average Contestant model. The predicted outcomes were "ex-post" (after the fact) predictions, in that we needed data from the live games to set certain simulation parameter values, particularly relating to Watson's buzzability. We were encouraged to see that the predicted rates of Watson winning a game (62%), leading going into Final Jeopardy (72%) and winning by lockout (27%) were within the standard error over 42 games of the actual rates (64%, 76%, and 26% respectively). There were more significant deviations on the low side in predicted final scores of Watson (15800 vs. 18400 actual) and of the humans (9300 vs. 10900 actual) but both were still within 95% confidence intervals.

By the start of the second series of Sparring Games, we were able to make "ex-ante" (before the fact) predictions, before Watson had actually played against human champions. These predictions were based mostly on using J! Archive data to tune parameters of the Champion model, as well as semi-educated guesses regarding the improvement in buzzability of human champions, and how aggressively they would seek out and wager on the Daily Doubles. The actual vs. predicted statistics are reported below in Table 2. Most of the ex-ante simulation stats turned out to be remarkably close to the actual results; only the rates of Watson leading in FJ, Watson's board control (i.e., how often Watson selected the next square) and human lockout rate differed by more than one standard error.

We then examined how much improvement could be obtained by ex-post recalibration of the Champion model, based on actual stats in the Sparring Games. As seen in Table 2, our best ex-post predictions failed to significantly improve on the ex-ante predictions. While there was notable improvement in Watson FJ lead and human lockout rates, the predictions of Watson lockout rate and human final score were noticeably worse.

| Statistic | Actual | Ex-ante sim | Ex-post sim |
|---|---|---|---|
| Watson win rate | $0.709 \pm 0.061$ | 0.724 | 0.718 |
| Watson lockout | $0.545 \pm 0.067$ | 0.502 | 0.493 |
| Watson FJ lead | $0.891 \pm 0.042$ | 0.806 | 0.830 |
| Watson board control | $0.500 \pm 0.009$ | 0.516 | 0.515 |
| Watson DDs found | $0.533 \pm 0.039$ | 0.520 | 0.517 |
| Watson final score | $23900 \pm 1900$ | 24950 | 24890 |
| Human final score | $12400 \pm 1000$ | 12630 | 13830 |
| Human lockout | $0.018 \pm 0.018$ | 0.038 | 0.023 |

Table 2: Comparison of actual mean statistics ($\pm$ std. error) in 55 Series 2 Sparring Games vs. ex-ante and ex-post predicted results in 30k simulation trials.

## 3. Optimizing Watson's Strategies Using the Simulation Model

The simulator described in the previous section enables us to estimate Watson's performance for a given set of candidate strategy modules, by running extensive contests between a simulation model of Watson and two simulated human opponents. The Watson stochastic process models use the same performance metrics (i.e., average attempt rate, precision, DD and FJ accuracies) as in the human models. The parameter values were estimated from J! Archive test sets, and were updated numerous times as Watson improved over the course of the project. The Watson model also estimates buzzability, i.e., its likelihood to win the buzz against humans of various ability levels. These estimates were initially based on informal live demo games against IBM Researchers, and were subsequently refined based on Watson's performance in the Sparring Games. We estimated Watson's buzzability against two humans at ∼80% for average contestants, 73% for Champions, and 70% for Grand Champions.

Computation speed was an important factor in designing strategy modules, since wagering, square selection and buzz-in decisions need to be made in just a few seconds. Also, strategy runs on Watson's "front-end," a single server with just a few cores, as its 3000-core "back-end" was dedicated to QA computations. As a result, most of Watson's strategy modules run fast enough so that hundreds of thousands of simulated games can be performed in just a few CPU hours. This provides a solid foundation for evaluating and optimizing the individual strategy components, which are presented below. Some strategy components (endgame buzz threshold, endgame DD betting, and Game 2 FJ betting) are based on compute-intensive Monte-Carlo trials; these are too slow to perform extensive offline evaluation. Instead, these strategies perform live online optimization of a single strategy decision in a specific game state.

### 3.1 Daily Double Wagering

We implemented a principled approach to DD betting, based on estimating Watson's likelihood of answering the DD clue correctly, and estimating how a given bet will impact Watson's overall winning chances if he gets the DD right or wrong. The former estimate is provided by an "in-category DD confidence" model. Based on thousands of tests on historical categories containing DDs, the model estimates Watson's DD accuracy given the number of previously seen clues in the category that Watson got right and wrong.

To estimate impact of a bet on winning chances, we follow the work of Tesauro (1995) in using Reinforcement Learning (Sutton & Barto, 1998) to train a Game State Evaluator (GSE) over the course of millions of simulated Watson-vs-humans games. Given a feature-vector description of a current game state, the GSE implements smooth nonlinear function approximation using a Multi-Layer Perceptron (Rumelhart, Hinton, & Williams, 1987) neural network architecture, and outputs an estimate of the probability that Watson will ultimately win from the current game state. The feature vector encoded information such as the scores of the three players, and various measures of the remaining amount of play in the game (number of remaining DDs, number of remaining clues, total dollar value of remaining clues, etc.).

The combination of GSE with in-category confidence enables us to estimate $E(bet)$, i.e. the "equity" (expected winning chances) of a bet, according to:

$$E(bet) = p_{DD} * V(S_W + bet, ...) + (1 - p_{DD}) * V(S_W - bet, ...) \qquad (1)$$

where $p_{DD}$ is the in-category confidence, $S_W$ is Watson's current score, and $V()$ is the game state evaluation after the DD has been played, and Watson's score either increases or decreases by the bet. We can then obtain an optimal risk-neutral bet by evaluating $E(bet)$ for every legal bet, and selecting the bet with highest equity. During the Sparring Games, our algorithm only evaluated round-number bets (i.e., integer multiples of $100), due both to computational cost as well as the possibility to obtain extra winning chances via a first-place tie or a "lock-tie" scenario as described in section 1.1. For the Exhibition Match, tie finishes were not possible, and we had sped up the code to enable evaluation of non-round wagers. This accounted for the strange wager values that were the subject of much discussion in the press and among viewers.

In practice, a literal implementation of risk-neutral betting according to Equation 1 takes on a frightening amount of risk, and furthermore, the calculation may contain at least three different sources of error: (i) the GSE may exhibit function approximation errors; (ii) the simulator used to train the GSE may exhibit modeling errors; (iii) confidence estimates may have errors due to limited test-set data. We therefore chose to adjust the risk-neutral analysis according to two established techniques in Risk Analytics. First, we added a penalty term to Equation 1 proportional to a bet's "volatility" (i.e., standard deviation over right/wrong outcomes). Second, we imposed an absolute limit on the allowable "downside risk" of a bet, defined as the equity difference between the minimum bet and the actual bet after getting the DD wrong. Due to function approximator bias, the latter technique actually improved expectation in some cases, in addition to reducing risk. We observed this in certain endgame states where the neural net was systematically betting too much, due to underestimation of lockout potential.

The overall impact of risk mitigation was a nearly one-third reduction in average risk of an individual DD bet (from 16.4% to 11.3%), at the cost of reducing expected winning chances over an entire game by only 0.3%. Given that Watson finds on average ~1.5-2.0 DDs per game (depending on how aggressively the opponents also seek DDs), this implies that the equity cost per DD bet of risk mitigation is quite small, and we regarded the overall tradeoff as highly favorable.

### 3.1.1 Illustrative Example

Figure 6 illustrates how the DD bet analysis operates, and how the resulting bet depends strongly on in-category confidence. The example is taken from one of the Sparring Games, where Watson got four consecutive clues right in the first category at the start of Double Jeopardy, and then found the first DD in attempting to finish the category. At this point, Watson's score was 11000 and the humans each had 4200. Watson's in-category confidence took its maximum value, 75%, based on having gotten four out of four correct answers previously in the category. Watson chose to wager $6700, which is a highly aggressive bet by human standards. (Fortunately, he got the DD clue right!)

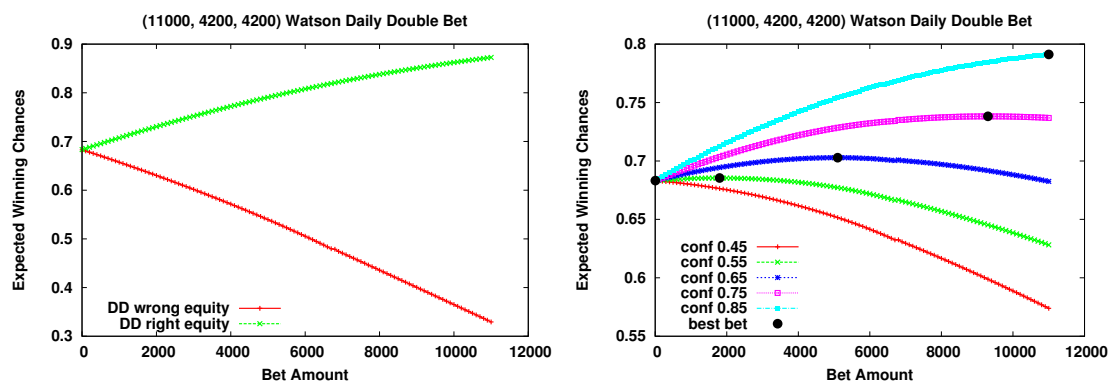

Figure 6: (left) Equity estimates getting the DD right (top curve) and wrong (bottom curve). (right) Bet equity curves at five differences in-category confidence levels, from 45% to 85%. Black dots show how the optimal risk-neutral bet increases with confidence.

The left figure shows neural net equity estimates for getting the DD right (top curve) and wrong (bottom curve) at various bet amounts. These curves are extremely smooth with gently decreasing slopes. The right plot shows the resulting equity-vs-bet curve at Watson's actual 75% confidence level (magenta curve), along with four other curves at different confidence values ranging from 45% to 85%. Black dots on each curve indicate the best risk-neutral bet, and we can see how the bet steadily increases with confidence, from $5 at 45%, to approximately $9300 at the actual 75%, and finally to the entire $11000 at a (hypothetical) confidence of 85%.

We also note the effect of risk mitigation, which reduced Watson's actual bet from $9300 to $6700. According to extensive Monte-Carlo analysis of this bet, risk mitigation reduced Watson's equity by only 0.2% (from 76.6% to 76.4%), but it entailed significantly less downside risk (more than 10%) in the event that Watson got the DD wrong. With a protection-to-cost ratio of over 50 to 1, we consider risk mitigation to have provided in this case an inexpensive form of disaster insurance, and the Watson team members were relieved to see that Watson did not risk his lead on this DD bet.

### 3.1.2 Endgame Monte-Carlo Wagers

For the Series 2 Sparring Games, we significantly boosted the simulation speed for regular clues and Final Jeopardy. This enabled replacement of neural net wagering in endgame states by a routine based on live Monte-Carlo trials. This analysis gives essentially perfect knowledge of which bet achieves the highest win rate in simulation, although it is still subject to modeling errors and confidence estimation errors. It also eliminated the primary weakness in Watson's DD strategy, as neural net misevaluations in endgames often resulted in serious errors that could considerably exceed 1% equity loss. As detailed below in section 3.1.3, usage of Monte-Carlo analysis for endgame wagers yielded a quite significant reduction (more than a factor of two) in Watson's overall error rate in DD betting.

With few clues remaining before Final Jeopardy, the dependence of equity on a player's score can exhibit complex behavior and discontinuities, in contrast to the smooth monotone behavior observed in early and mid-game states. A striking example is plotted in Figure 7. This was an endgame DD bet from the Series 2 Sparring Games where Watson had 19800, the humans had 13000 and 14300, and there were only four remaining clues (two $400 and two $800). Watson was 4-for-4 in the category, which translated into 71.8% DD confidence. (We opted for a more conservative estimate than the 75% figure mentioned earlier, due to possible confidence estimation errors.)

We see on the left that the right/wrong equity curves exhibit complex acceleration and deceleration, as well as periodic jumps with periodicity of $400. These may reflect scores where a discrete change occurs in the combinations of remaining squares needed to reach certain FJ breakpoints, such as a lockout. The equity-vs-bet curve on the right also displays interesting multi-modal behavior. There is a peak "lead-preserving bet" around $3000. At $6400, the curve begins a steep ascent – this is the point at which a lockout becomes mathematically possible. The curve continues to rise until about $12000, where a correct answer assures the lockout, and then it falls off.

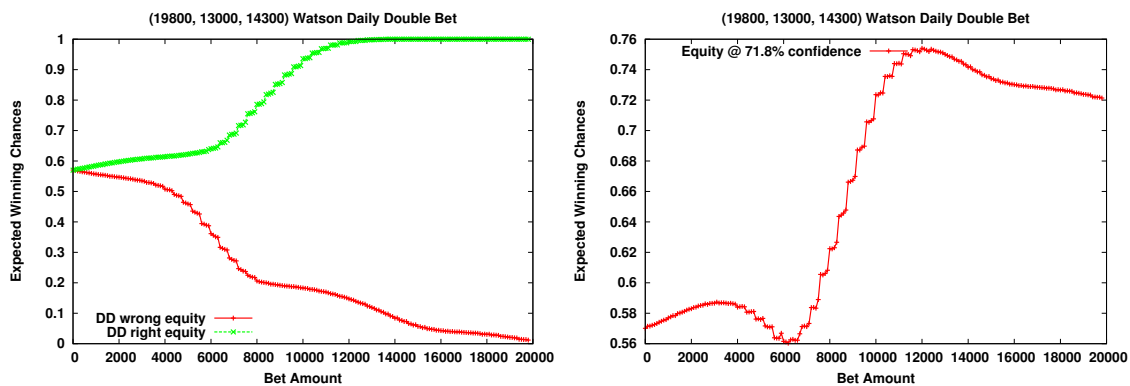

Figure 7: MC DD bet analysis. (left) Equity estimates getting the DD right (top curve) and wrong (bottom curve). (right) Bet equity curve at Watson's estimated in-category confidence of 71.8%.

### 3.1.3 Performance Metrics and Error Analysis

We assessed the performance of neural net DD wagering by two different methods. First, we noted an improved win rate in simulations when compared to Watson's previous DD wagering algorithm, a set of heuristic betting rules that tried to safely add to Watson's lead or safely catch up, without dropping below certain strategically important score breakpoints. While the heuristic rules embodied sound logic, they suffered a major limitation of not taking Watson's in-category confidence into account, so that they would generate the same wager regardless of confidence.

Using the heuristic DD betting rules, Watson's simulated win rate was 61%. With neural net DD wagering using a default confidence value for every DD, the win rate improved to 64%. When we added emulation of live DD confidence values to the simulation, the result was a further jump in win rate, to 67%. We regarded this as a quite significant performance improvement, given that the DD betting algorithm is only used about 1.5-2.0 times per game.

The second performance metric utilizes extensive offline Monte-Carlo analysis of many neural network bets to estimate average "equity loss" per DD bet, i.e., the average difference between the equity of the true best bet, with highest simulated win rate, and the equity of the bet selected by the neural net. This figure was approximately 0.6% per DD bet, which was quite good, as it implied that the overhead to improve Watson's win rate via improved DD wagering was less than 1.2%. Most of the equity loss was due to large errors in endgame states. As mentioned earlier, we substantially reduced the loss rate by implementing DD wagering in endgames based on live Monte-Carlo simulations. This reduced Watson's average equity loss per DD bet from 0.6% to ∼0.25%, with about half of that loss rate resulting from deliberately imposed risk mitigation. Hence we were satisfied that Watson's DD algorithm was close enough to optimal for all practical purposes.

### 3.1.4 Human DD Error Analysis

An interesting and nontrivial question arising from the above analysis is how does Watson's equity loss rate in DD wagering compare with that of human contestants. The main difficulty in attempting such analysis is that contestants' confidence in answering the DD clue correctly is largely unobservable in historical data. We have no way to know their confidence in the category, and their knowledge on previous clues in the category is only revealed if they won the buzz and gave an answer. In the absence of confidence information, it is hard to ascribe an error level to any individual DD bet, although we may be able to assess average wagering error over an entire population of contestants.

We devised two methods that respectively should provide lower and upper bounds on population average wagering error, given sufficient samples of historical DD bets. The first method is a historical replacement technique similar to those presented in sections 2.3 and 3.2. For each historical DD, we first use the Average Contestant simulator to run many trials starting from the actual outcome state of the DD, reflecting the contestant's actual bet and actual right or wrong answer. We then replace the contestant's bet with some algorithmically computed bet, wherein confidence is estimated solely from observable information, and we rerun trials from the modified outcome state, where the contestant's score is changed due to the change of bet. Averaged over many DDs, the equity difference between human bets and algorithmic bets should indicate which approach is better. If the algorithmic bets prove to be better, the equity difference should provide a lower bound on the true human error rate: since the algorithm does not have access to private confidence information, it would presumably obtain even better results given such information.

There could be issues with such an approach in faithfully simulating what would have happened if the contestant had bet differently, as changing the bet might have changed subsequent square selection (DD seeking) and subsequent DD wagering. We minimized these issues by limiting the analysis to last-DD situations. Our historical dataset contained more than 2200 regular episodes where both DDs were played in the second round. By analyzing last-DD states in these episodes, we avoid having to simulate subsequent DD wagers, and the effect of subsequent square selection should be minimal since there are no more DDs to be found. Also, the last-DD states allow us to use endgame Monte-Carlo for algorithmic wagering, which should give stronger results than neural net wagering. Confidence estimates in the Monte-Carlo calculation were based on the historical mean accuracy on second-round DDs given the row location. This ranges from $\sim$72% for the top two rows to $\sim$57% for the bottom row.

As seen in Table 3, results of the analysis showed that contestant bets on average obtained about 2.9% less equity per bet than Monte-Carlo algorithmic bets. As this should constitute a lower bound on the true human error rate in last-DD states, whereas WATSON's error rate is 0.25% overall and near-perfect in endgames states, this provides compelling evidence of WATSON's superiority to human contestants, at least for last-DD wagers.

Our second analysis method ignores the actual right/wrong contestant answers, and instead uses the Monte-Carlo analysis to calculate the equity loss of a human bet, assuming that the row-based mean DD accuracies provide "correct" confidence estimates. This type of analysis should overestimate human errors, as it unduly penalizes small bets based on low private confidence, and large bets based on high private confidence. Results of this analysis on the last-DD dataset show an average error rate of 4.0% per DD bet. This result is consistent with an estimated lower bound on the error rate of 2.9%, and in combination, both results provide reasonable evidence that the actual human error rate lies in between the estimated bounds.

| Last-DD Player | Lower Bound | Upper Bound | Avg. Human Bet | Avg. MC Bet |
|---|---|---|---|---|
| All | 2.9% | 4.0% | $2870 | $6220 |
| A (leader) | 0.9% | 2.2% | $2590 | $5380 |
| B,C (trailer) | 4.8% | 5.5% | $3120 | $6970 |

Table 3: Lower and upper bounds on average equity loss rates of historical human contestant wagers on the last DD of the game. The average human bets vs. recommended MC bets (at default confidence values) are also displayed.

A closer examination of the analysis reveals that humans systematically wager these DDs far too conservatively. After segregating the data according to whether the DD player is leading or trailing, we found that this conservatism is manifest in both cases. Leading players on average bet $2590, whereas the average recommended MC bet is $5380. For trailing players, the average bet is $3120 vs. an average recommended bet of $6970. A more startling finding is that these errors are far more costly to trailers than to leaders, in terms of equity loss. The lower and upper bounds on error rate for leaders are 0.9% and 2.2%, while for trailers, the respective bounds are 4.8% and 5.5%! We further discuss the implications of these results for human contestants in section 4.2.

### 3.1.5 Multi-game DD Wagering

As mentioned in section 2.6, Game 1 and Game 2 of our Exhibition Match required distinct wagering strategies, with both differing from single-game wagering. We trained separate neural networks for Game 1 and Game 2. The Game 2 net was trained first, using a plausible artificial distribution of Game 1 final scores.

Having trained the Game 2 neural net, we could then estimate the expected probabilities of Watson ending the match in first, second, or third place, starting from any combination of Game 1 final scores, by extensive offline Monte-Carlo simulations. We used this to create three lookup tables, for the cases where Watson ends Game 1 in first, second, or third place, of Watson match equities at various Game 1 final score combinations, ranging from (0, 0, 0) to (72000, 72000, 0) in increments of 6000. (Since adding or subtracting a constant from all Game 1 scores has no effect on match equities, we can without loss of generality subtract a constant so that the lowest Game 1 score is zero.) Since match equities are extremely smooth over these grid points, bilinear interpolation provides a fast and highly accurate evaluation of Game 1 end states. Such lookup tables then enabled fast training of a Game 1 neural net, using simulated matches that only played to the end of Game 1, and then assigned expected match-equity rewards using the tables.

Based on our earlier experience, we added a heuristic "lockout-potential" feature to the Game 2 input representation, using a heuristic sigmoidal formula to estimate the probability that Watson would win the match by lockout, given the Game 1 and Game 2 scores and the dollar value of remaining clues. This feature appeared to enable highly accurate Game 2 evaluations and eliminated the large endgame equity losses that we had observed in using the single-game neural net.

An important and difficult new issue we faced in the Exhibition Match format was how to assign relative utilities to finishing in first, second and third place. Unlike the Sparring Games where the only reasonable objective was to finish first, the team extensively debated how much partial credit should be ascribed to a second-place finish, in which Watson defeated one of the two greatest *Jeopardy!* contestants of all time. Ultimately we decided to base the match DD wagering on full credit for first, half credit for second, and zero credit for a third place finish. Such an objective acts as an additional type of risk control, where Watson would only prefer a large bet over a smaller and safer bet if the large bet's upside (increased chances to win) exceeded its downside (increased chances of finishing third). There was unanimous team consensus that such a risk would always be worth taking.

Defining Watson's match equity as probability of finishing first plus 0.5 times probability of finishing second, we estimated Watson's average match equity loss per DD bet at about 0.3% in Game 1 and 0.55% in Game 2. The majority of loss in each case was due to risk controls.

## 3.2 Final Jeopardy Wagering

Our approach to Final Jeopardy wagering involves computation of a "Best-Response" strategy (Fudenberg & Tirole, 1991) (a standard game-theoretic concept) to the human FJ model presented in section 2.3. We considered attempting to compute a Nash Equilibrium strategy (Fudenberg & Tirole, 1991), but decided against it for two reasons. First, due to the imperfect information in Final Jeopardy (contestants know their own confidence given the category title, but do not know the opponents' confidence), we would in principle need to compute a Bayes-Nash equilibrium (BNE), which entails considerably more modeling and computational challenges. Second, it seems far-fetched to assume that Watson's opponents would play their part of a Nash equilibrium or BNE, since average contestants have not studied game theory.

By using a Best-Response instead of a BNE strategy (assuming we could calculate it), we aim to more effectively exploit typical human wagering "errors" as recorded in the historical data. We realized that this potentially exposes Watson to two types of risk. First, if Watson's live opponents turned out to use BNE or other more sophisticated strategies than those built into our human models, Watson might do better by playing the BNE strategy. We judged this risk to be sufficiently rare that Watson would surely do better over the course of many Sparring Games by simply playing the Best-Response. Second, since the Best-Response is essentially a deterministic strategy, a contestant who observed Watson play many Final Jeopardy rounds might be able to detect and optimally exploit Watson's wagering strategy. However, such observations were limited by the procedures for conducting the Sparring Games, as any contestant would only play 2-3 games against Watson, and would observe only another 1-2 games as a spectator. In some situations, we were additionally able to randomize Watson's bet over a range of bets with approximately equal expected outcomes; this made it more difficult for humans to infer Watson's wagering logic from a limited number of observations.

Computation of the Best-Response proceeds as follows. First, we consult an "FJ prior accuracy" regression model to estimate Watson's confidence given the category title. This model was trained on samples of Watson's performance in thousands of historical FJ categories, using NLP-based feature vector representations of the titles. Second, given Watson's confidence and the human accuracy/correlation parameters, we derive analytic probabilities of the eight possible right/wrong outcomes. Third, for a given FJ score combination, we draw on the order of 10000 Monte-Carlo samples of bets from the human models. Finally, we evaluate the equity of every legal bet, given the human bets and the right/wrong outcome probabilities, and select the bet with highest equity.

Our initial implementation of the above algorithm was too slow to use in live play. Fortunately, after extensive offline analysis using Watson's default confidence, we discovered that the Best-Response output could be expressed in terms of a fairly simple set of logical betting rules. For example, one of the rules for B stipulates:

> If B has at least two-thirds of A, and B has less than 2C, check whether 2C-B (the amount to cover C's doubled score) is less than or equal to 3B-2A (the maximum two-thirds bet). If so, then bet 2C-B, otherwise bet everything.

Thus in the Series 1 Sparring Games, we deployed the rule-based encapsulation of the Best-Response calculation, but with one specific exception where Watson is A, and the B player may reasonably consider a so-called "two-thirds" bet, i.e., a small bet that aims to guarantee a win whenever A is wrong, assuming A makes the standard shut-out bet (2B-A+1). In some of these situations, the Best-Response calculation calls for WATSON to counter B by making a small, tricky "anti-two-thirds" bet. Whether this really wins more often depends on exact model parameter values, which have considerable uncertainty, as discussed earlier in section 2.3. Moreover, if WATSON bets small and gets FJ wrong, it suggests shakiness in his ability to play Final Jeopardy, which might be exploited in subsequent games, as well as generally looking bad. Conversely, if WATSON bets small and gets FJ right, it risks an embarrassing loss where WATSON could have won but failed to bet enough to win. For all these reasons, the team preferred to override the Best-Response calculation, and to have WATSON simply make the standard bet in this scenario. This proved to be a judicious choice in hindsight, as there were five games in Series 1 where the Best-Response would have bet anti-two-thirds, but B did not bet two-thirds in any of those games.

For the Series 2 Sparring Games, the Best-Response computation had been sped up enough to enable live wager computations. The team continued to debate whether to allow anti-two-thirds bets, and ultimately decided it was worth the risk if WATSON had unusually low confidence (as happens, for example, in categories such as US Presidents, Shakespeare, and US Cities). As it turned out, there were only two games in this series where the Best-Response would have bet anti-two-thirds. B bet two-thirds in one game but not in the other game. WATSON did not have low confidence in either FJ category, so there were no live test results of the anti-two-thirds strategy.

We assessed the performance of the Best-Response strategy via the historical replacement technique presented in section 2.3; results are displayed in Table 4. The first column gives the actual human win rates in the A, B, C roles. The second column shows win rates for the constrained Best-Response deployed in the Series 1 Sparring Games, in which A always bets to cover 2B. Note that this algorithm considerably improves over actual human results as B and C, and provides a smaller but noticeable improvement over human A bets. We attribute the latter gain partly to more consistent betting, and partly to judicious bets in some cases to tie 2B, rather than trying to surpass 2B by $1. The last column gives results for the full Best-Response algorithm including general Best-Response bets as A. There is nearly a 1% improvement in A's win rate over the constrained Best-Response; this provides support for the efficacy of anti-two-thirds and other sophisticated strategies, but is not quite statistically significant at 2092 trials.

|   | Human | Constrained Best-Response | Full Best-Response |
|---|-------|---------------------------|--------------------|
| A | 65.3% | 67.0% | 67.9% |
| B | 28.2% | 34.4% | 34.4% |
| C | 7.5%  | 10.5% | 10.5% |

Table 4: Comparison of actual human win rates with win rates of the constrained and full Best-Response strategies, by historical replacement in 2092 non-locked FJ situations from past episodes.

For the Exhibition Match, we devised live Best-Response algorithms for Game 1 and Game 2 based on Monte-Carlo samples of the human betting models of section 2.6, and probabilities of the eight right/wrong outcomes given Watson's FJ category confidence. For the first-game FJ, we can't evaluate directly from the FJ outcomes since there is still a second game to play. The evaluation is instead based on interpolation over the lookup tables discussed in section 3.1.5 denoting Watson's match equities from various first-game score combinations.

Due to modeling uncertainties in Game 2 FJ, we devoted much effort to interpreting the Best-Response output in terms of logical betting rules, as well as deciding whether any Best-Response decisions should be overridden. We ultimately decided to allow Watson to venture an anti-two-thirds bet as A only if the predicted category confidence was unusually low; otherwise Watson would always bet to guarantee a win as A by answering correctly. For wagering as B, the betting rules would only attempt to finish ahead of A if it did not diminish Watson's chances of finishing ahead of C. This naturally emerged from the match utility function which assigned half-credit for a second place finish. Finally, for wagering as C, the Best-Response output was too complex to derive human-interpretable rules, so Watson was prepared to run the live calculation in this case. As it turned out, all of the above work was superfluous, since Watson had a lockout in Game 2 of the Exhibition Match.

## 3.3 Square Selection

We considered four different factors that could conceivably be relevant to the optimal overall objective for Watson in deciding which square to select in a given game state:

- Selecting a Daily Double square: Finding the DDs quickly can provide an excellent opportunity for Watson to significantly boost his game standing, while also denying that opportunity to the other players. The potential downside is that Watson may only have a small bankroll to wager, and may have little or no evidence as to assess his likelihood of answering the DD clue correctly.

- Retaining control of the board: this involves estimating categories and/or square values where Watson has the greatest chance to win the buzz and answer correctly. This would give Watson another chance to try to find a DD, if the selected square turns out to be a regular clue.

- Learning the "essence" of a category, i.e., gathering information about the category such as the type of correct answers, so as to improve accuracy on subsequent clues in the category (Prager et al., 2012). This consideration would suggest selecting low-value squares first, so that accuracy would be improved on higher-value squares.

- Maximizing expected score change: This concept seeks the best combination of high expected accuracy with highest dollar value of available clues to obtain the biggest boost in Watson's score on the next square.

We used the simulator to systematically investigate numerous weighted combination of the above four factors. These studies were performed using Champion and Grand Champion human models, which featured overt DD seeking, aggressive DD wagering, and high

DD accuracy. Our results showed that, prior to all DDs being revealed, finding DDs is overwhelmingly the top factor in maximizing WATSON's win rate, and retaining control is second in importance. Learning the essence of a category appears to provide an effective strategy only after all DDs have been found, and maximizing expected score change did not appear to be useful in improving WATSON's win rate.

These findings led us to deploy an algorithm that selects squares as follows. First, if there are any unrevealed DDs, a square $i^*$ is selected that maximizes $p_{DD}(i) + \alpha p_{RC}(i)$ where $p_{DD}(i)$ is the probability that square $i$ contains a DD, $p_{RC}(i)$ is an estimated probability that WATSON will retain control of the board if $i$ does not contain a DD, and $\alpha = 0.1$ yielded the best win rate. The first term is calculated using Bayesian inference, as described below in section 3.3.1. The second probability is estimated by combining the simulation model of human performance on regular clues with a model of WATSON that adjusts its attempt rate, precision and buzzability as function of number of right/wrong answers previously given in the category. Second, after all DDs in the round have been found, the algorithm then switches to selecting the lowest dollar value in the category with the greatest potential for learning about the category: this is based on the number of unrevealed clues in the category and their total dollar value.

### 3.3.1 BAYESIAN DD PROBABILITY CALCULATION

We calculate $p_{DD}(i)$, the probability that square $i$ contains a DD, according to principles of Bayesian inference: we combine Bayesian prior probabilities, taken from historical frequencies of DD locations, with evidence from revealed questions according to Bayes' rule, to obtain posterior probabilities. The computation is easy to perform incrementally as each individual question is revealed, and works somewhat differently in Round 1 than in Round 2, due to the different number of available DDs.

In Round 1, there is only one DD, so computation of posterior probabilities is easy. Let $p(i)$ denote the prior probability that square $i$ contains a DD. Let $p(\neg i) = 1 - p(i)$ denote the prior probability that $i$ does not contain a DD. Now assume that a square $j$ is revealed not to contain a DD. The posterior probability $p(i|\neg j)$ according to Bayes' rule is given by:

$$p(i|\neg j) = \frac{p(\neg j|i)p(i)}{p(\neg j)} \qquad (2)$$

where $p(\neg j|i) = 1$ by definition, assuming $i \neq j$. Of course, once the DD has been revealed, all $p(i)$ values for other squares are set to zero.

In Round 2, there are two DDs, and their probabilities are not independent, since both DDs cannot be located in the same column, plus there are column pair frequencies in the historical data that may not be explainable by an independent placement model. We therefore maintain a joint probability distribution $p(i, j)$ indicating the probability that squares $i$ and $j$ both contain DDs. We initialize $p(i, j)$ to prior values, using joint DD location frequencies in the historical data. Now assume that square $k$ is revealed not to contain a DD. The posterior probability $p(i, j|\neg k)$ is computed according to Bayes' rule as:

$$p(i, j|\neg k) = \frac{p(\neg k|i, j)p(i, j)}{p(\neg k)} \qquad (3)$$

where $p(\neg k) = 1 - p(k)$ is the marginal distribution of a single DD, integrating over possible locations of the second DD, and $p(\neg k | i, j) = 1$ if $k \neq i$ and $k \neq j$, else it equals 0. Note that the constraint of two DDs never appearing in the same column is enforced by setting the prior $p(i, j) = 0$ if squares $i$ and $j$ are in the same column. This guarantees that the posterior will always equal 0, since Bayes' rule performs multiplicative updates.

If a square $k$ is discovered to contain a DD, then the rest of the board can be updated similarly:

$$p(i, j | k) = \frac{p(k | i, j) p(i, j)}{p(k)} \tag{4}$$

where $p(k | i, j) = 1$ if $k = i$ or $k = j$, else it equals 0.

### 3.3.2 Square Selection Performance Metrics

| Live DD strategy | No live DD strategy | Win rate | DDs found | Board control |
|---|---|---|---|---|
| LRTB | LRTB | 0.621 | 0.322 | 0.512 |
| Simple DD seeking | LRTB | 0.686 | 0.510 | 0.518 |
| Bayes (max $p_{DD}$) | LRTB | 0.709 | 0.562 | 0.520 |
| Bayes (max $p_{DD}$) | Post-DD learning | 0.712 | 0.562 | 0.520 |
| max($p_{DD} + 0.1 p_{RC}$) | Post-DD learning | 0.714 | 0.562 | 0.520 |

Table 5: Simulation results in two-game matches vs. Grand Champions using various square selection strategies (500k trials). LRTB denotes left-to-right, top-to-bottom square selection.

In Table 5 we report on extensive benchmarking of Watson's performance using five different combinations of various square selection algorithms. The first column denotes strategy when there are available DDs to be played in the round, while the second column denotes strategy after all DDs in the round have been played. These experiments utilized the two-game match format with Grand Champion models of the human contestants. As stated earlier, these human models employ aggressive DD and FJ wagering, and simple DD seeking using the known row statistics when DDs are available. Simulations of Watson use right/wrong answers drawn from historical categories, so that Watson will exhibit learning from revealed answers in a category. As an interesting consequence of Watson's learning, we model human square selection with no remaining DDs according to an "anti-learning" strategy, intended to frustrate Watson's learning, by selecting at the bottom of the category with greatest potential benefit from learning. We actually observed this behavior in informal testing with very strong human players (Ed Toutant and David Sampugnaro) just before the Exhibition Match, and there was evidence that Jennings and Rutter may have selected some clues in the Exhibition based on this concept.

Results in Table 5 show that the weakest performance is obtained with an extremely simple baseline strategy of Left-to-Right, Top-to-Bottom (LRTB), i.e., always selecting the uppermost square in the leftmost available column, while our actual deployed strategy in the Exhibition gives the strongest performance. Consistent with all our earlier findings, we see in Table 5 that DD seeking is extremely important, especially when playing against strong humans that overtly seek DDs. Our Bayesian DD seeking method is significantly better than simple DD seeking based solely on the row frequencies of DDs. When Watson and humans all use simple DD seeking, Watson finds 51.0% of the DDs (roughly in line with its 51.8% average board control) and its match win rate is 68.6%. When Watson switches to Bayesian DD seeking, its rate of finding DDs jumps to 56.2%, even though board control is virtually unchanged at 52.0%, and its win rate increases by 2.3% to 70.9%. On the other hand, if Watson does no DD seeking, and simply uses Left-to-Right, Top-to-Bottom selection, its rate of finding DDs plunges to 32.2% and its overall win rate drops to 62.1%.

The additional effects of seeking to retain control of the board, and selecting categories with greatest potential for learning after all DDs are revealed, are smaller but statistically significant after 500k trials. We find that optimizing the weight on $p_{RC}$ increases win rate by 0.2%, and maximizing learning potential with no remaining DDs adds another 0.3% to Watson's win rate.

## 3.4 Confidence Threshold for Attempting to Buzz

Watson will attempt to buzz in if the confidence in its top-rated answer exceeds an adjustable threshold value. In the vast majority of game states, the threshold was set to a default value near 50%. While we did not have analysis indicating that this was an optimal threshold, we did have a strong argument that a 50% threshold would maximize Watson's expected score, which ought to be related to maximizing Watson's chance to win. Furthermore, it was clear that general default buzzing at 50% confidence was better than not buzzing, since Watson's expected score change (0) would be the same in both cases, but the opponents would have a much better chance to improve their scores if Watson did not buzz.

From an approximate threshold calculation based on a "Max-Delta" objective (described in Appendix 2), we had suggestive evidence that the initial buzz threshold should be more aggressive. Subsequent more exact Monte-Carlo analysis for endgames (Appendix 2) and for early game states (section 4.3) provides substantial backing for an aggressive initial threshold below 50%. Nevertheless, since Watson tended to be slightly overconfident in the vicinity of 50% nominal confidence, and since many of Watson's wrong answers in this vicinity clearly revealed the correct answer to the opponents, the 50% default threshold may have been a prudent choice.

Near the end of the game the optimal buzz threshold may vary significantly from the default value. One special-case modified buzz policy that we devised for endgames uses a "lockout-preserving" calculation. For Round 2 states with no remaining DDs, if Watson has a big lead, we calculate whether he has a guaranteed lockout by not buzzing on the current square. If so, and if the lockout is no longer guaranteed if Watson buzzes and is wrong, we prohibit Watson from buzzing, regardless of confidence.

In principle, there is an exact optimal binary buzz-in policy within our simulation model $\vec{B}^*(c, D) = (B_0^*(c, D), B_1^*(c, D), B_2^*(c, D), B_3^*(c, D))$ for any game state with a clue currently in play, given Watson's confidence $c$ and the dollar value $D$ of the current clue. The policy components $B_i^*(c, D)$ result from testing whether $c$ exceeds a set of optimal threshold values $\{\theta_i^*, i = 0, 1, 2, 3\}$. There are four such values corresponding to the four possible states in which Watson may buzz: the initial state, first rebound where human #1 answered incorrectly, first rebound where human #2 answered incorrectly, and the second rebound where both humans answered incorrectly. The optimal policy can be calculated using Dynamic Programming (DP) techniques (Bertsekas, 1995). This involves writing a recursion relation between the value of a current game state with $K$ clues remaining before FJ, and values of the possible successor states with $K - 1$ clues remaining:

$$V_K(s) = \int \rho(c) \sum_{j=1}^{5} p(D_j) \max_{\vec{B}(c, D_j)} \sum_{\delta} p(\delta | \vec{B}, c) V_{K-1}(s'(\delta, D_j)) dc \tag{5}$$

where $\rho(c)$ is the probability density of Watson's confidence, $p(D_j)$ denotes the probability that the next square selected will be in row $j$ with dollar value $D_j = \$400 * j$, the max operates over Watson's possible buzz/no-buzz decisions, $p(\delta | \vec{B}, c)$ denotes the probability of various unit score-change combinations $\delta$, and $s'$ denotes various possible successor states after the $D_j$ square has been played, and a score change combination $\delta$ occurred. (See Appendix 2 for a detailed discussion of how the recursion relation in Equation 5 is calculated.)

We implemented an exact DP solver which successively expands the root state to all successor states with $K - 1, K - 2, ..., 0$ clues remaining, where 0 denotes Final Jeopardy states. The FJ states are evaluated by Monte-Carlo trials, and values are propagated backward according to Equation 5 to ultimately compute the optimal buzz policy in the root node state. While this computation is exact within our modeling assumptions, it is too slow to use in live play if $K \geq 2$, due to very high branching factor in the search tree.

In order to achieve acceptable real-time computation taking at most $\sim$1-2 seconds, we therefore implemented an Approximate DP calculation in which Equation 5 is only used in the first step to evaluate $V_K$ in terms of $V_{K-1}$, and the $V_{K-1}$ values are then based on plain Monte-Carlo trials (Tesauro & Galperin, 1996; Ginsberg, 1999; Sheppard, 2002). Due to slowness of the exact DP calculation, we were unable to estimate accuracy of the approximate method for $K > 5$. However, we did verify that Approximate DP usually gave quite good threshold estimates (within $\sim$5% of the exact value) for $K \leq 5$ remaining squares, so this was our switchover point to invoke Approximate DP as deployed in the live Series 2 Sparring Games against human champions. An analogous algorithm based on match equities was also deployed in Game 2 of the Exhibition Match, but was indifferent on the final five clues in the live game, since Watson had a guaranteed win after either buzzing or not buzzing.

### 3.4.1 Illustrative Examples

The Approximate DP buzz-in algorithm easily handles, for example, a so-called "desperation buzz" on the last clue, where Watson must buzz in and answer correctly to avoid being locked out (e.g., suppose Watson has 4000, the human contestants have 10000 and 2000, and the final clue value is $1200). Generally speaking, optimal endgame buzzing shows the greatest deviation from default buzzing near certain critical score breakpoints, such as the crossover from third to second place, or from second to first place. When a player's score is just below one of these breakpoints, aggressive buzzing is usually correct. Conversely, with

a score just above a critical breakpoint, players should buzz much more conservatively, to guard against dropping below the breakpoint.

The most critical breakpoint is where a contestant achieves a guaranteed lockout. In near-lockout situations, the algorithm may generate spectacular movements of the buzz threshold that are hard to believe on first glance, but which can be appreciated after detailed analysis. An example taken from the Sparring Games is a last-clue situation where WATSON had 28000, the humans had 13500 and 12800, and the clue value was $800. The (initially) surprising result is that the optimal buzz threshold drops all the way to zero! This is because after buzzing and answering incorrectly, WATSON is no worse off than after not buzzing. In either case, the human B player must buzz and answer correctly in order to avoid the lockout. On the other hand, buzzing and answering correctly secures the win for WATSON, so this is a risk-free chance to try to buzz and win the game.

A more complex example occurred in a later game, where there were two squares remaining (the current one was $1200 and the final one was $2000), and WATSON had 31600, vs. 13000 and 6600 for the humans. Once again, any correct answer by WATSON wins the game. The analysis shows that WATSON can buzz regardless of confidence on both this clue and the next clue, and do just as well or better than not buzzing. On the first clue, if WATSON buzzes and is wrong, B needs to buzz and answer correctly, to reach a score of 14200, otherwise WATSON has a lockout at 30400. Now suppose WATSON also gets the second clue wrong, dropping to 28400. The score pair (28400, 14200) is now just as good for WATSON as the state (31600, 14200) if WATSON did not attempt either clue. In both cases, WATSON has a guaranteed win unless B answers correctly. In fact, if B is alert, she might deliberately not answer at (28400, 14200) as it is a "lock-tie" situation; this is actually better for WATSON than (31600, 14200), although our simulator does not model such behavior.

A critical example of the converse situation, where B's buzz-in threshold is much higher than normal, occurred in an earlier game. On the final clue ($2000 value) WATSON (A) had 25200, B had 12800, and C had 2600, and WATSON answered incorrectly on the initial buzz, dropping to 23200. Despite being a *Jeopardy!* champion, B unfortunately buzzed on the rebound and answered incorrectly, thereby locking himself out. Our analysis shows that the rebound buzz is a massive blunder (although understandable in the heat of live play): it offers no improvement in FJ chances if B is right, and forfeits any chance to win if B is wrong. If the roles were reversed, WATSON would have buzzed fairly aggressively on the initial buzz, to prevent A from achieving the lockout, but never would have buzzed on the rebound.

Finally, Figure 8 presents a $2000 last-square situation where we fix the human scores at (13000, 6600) and systematically study how WATSON's initial buzz threshold varies with score. There are several examples of huge threshold changes as breakpoints are crossed. For example, WATSON's threshold goes from very aggressive (0.12) just below 13000, to fairly conservative (0.65) just above 13000. There is additional complex behavior arising from specific score combinations involving all three players. For example, at 6400 WATSON can take extra risk of answering incorrectly, due to the chance that A may also answer incorrectly. This creates a situation where A=B+C and WATSON has extra chances to achieve a tie for first place. The same principle applies at 10400, where A=B+C arises if WATSON is wrong and A is right. A different combination comes into play at 10800 where

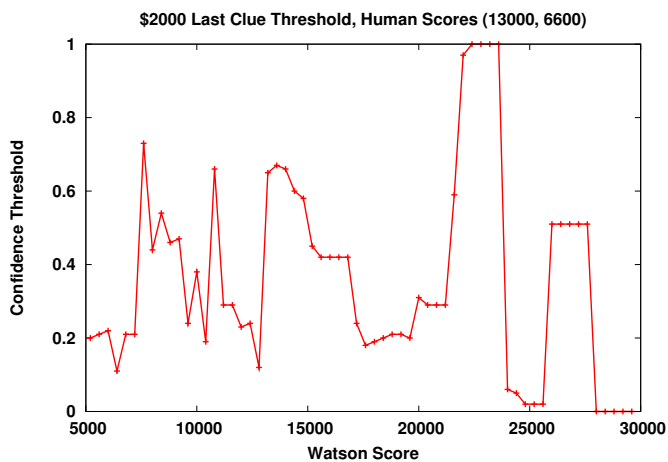

Figure 8: Watson's initial buzz threshold vs. score on the last clue ($2000) before FJ.

Watson has extra incentive not to buzz: if either A or C are right, Watson can bet to cover 2C and it will still constitute a two-thirds bet.

Figure 8 also shows huge swings where Watson is close to achieving a lockout. At 23000, Watson will never buzz, since there is no chance to get a lockout. At 25000, Watson has a free shot to try for the lockout, as discussed earlier. At 27000, Watson has a provisional lockout but needs to take some risk to block a correct answer by B, and at 29000, Watson has a free shot to prevent B from answering.

## 4. Lessons for Human Contestants

Now that Watson has retired as a *Jeopardy!* contestant, any future impact of our work in improving *Jeopardy!* performance will relate specifically to human contestants. In this section, we present a number of interesting insights that may help future contestants improve their overall winning chances.

### 4.1 Basics of Final Jeopardy Strategy

The observed FJ wagers in our J! Archive dataset suggest that many contestants appearing on the show devote scant effort to learning good strategies for FJ wagering, apart from the elementary concept of A wagering at least 2B-A to cover B's doubled score. While we don't intend in this section to provide a definitive treatise on FJ strategy, we can illustrate what we found to be the most important regions and separating boundaries in FJ strategy space in a single plot, shown below in Figure 9.

Since FJ scenarios are scale invariant, any scenario is uniquely determined by two variables: B's score relative to A, and C's score relative to B. The ratio of B to A is the most important quantity in Final Jeopardy, and the most important breakpoint (apart from B<A/2 which is a lockout) is B=2A/3, illustrated by the solid red line. All contestants should be familiar with the implications of this scenario, which is analogous to the game of

235

Matching Pennies, where A wins if the pennies match, and B wins if the pennies mismatch. If B has at least two-thirds of A, B can secure a win whenever A is wrong by making a small "two-thirds" bet ≤(3B-2A), assuming that A bets to cover 2B. However, this strategy is vulnerable to A making a small "anti-two-thirds" bet, which would give B no chance to win. Conversely, A's anti-two-thirds bet is vulnerable to B making a large bet to overtake A. A related breakpoint is the case where B≥3A/4: in this situation B's two-thirds bet can overtake A, so that A's anti-two-thirds option is eliminated.

Other important breakpoints include C=B/2 (green line): below this line, B can keep out C with a small bet ≤(B-2C), while above the line, B needs to bet at least (2C-B) to cover C's doubled score. The latter case may lead to a dilemma if (2C-B) exceeds the maximum two-thirds bet (3B-2A). The demarcation where this dilemma occurs is the magenta curve (2B=A+C), which is also known as the "equal spacing" breakpoint, since A-B=B-C.

Breakpoints that primarily affect C are the curves C=(A-B) (dark orange) and C=2(A-B) (gray). Basically, C needs to be able to reach the 2(A-B) curve to have any chance to win, so that C has no chance below the (A-B) curve. For scenarios lying between the two curves, C has a "minimum rational" bet of at least 2(A-B)-C, although a larger bet may be reasonable, for example, if C≥A/2 (dotted black curve) and can overtake A. Such a scenario would also dissuade A from trying an anti-two-thirds bet against B.

When C has at least 2(A-B), the general rule is to bet small enough to stay above this value. An additional upper bound emerging from our Best Response calculation occurs when C≥2B/3 (blue line), in cases where B had more than 3A/4. In this case, B has an incentive to bet to cover 2C, so that C has an opportunity to execute a two-thirds bet against B, which may yield more wins than simply staying above 2(A-B).

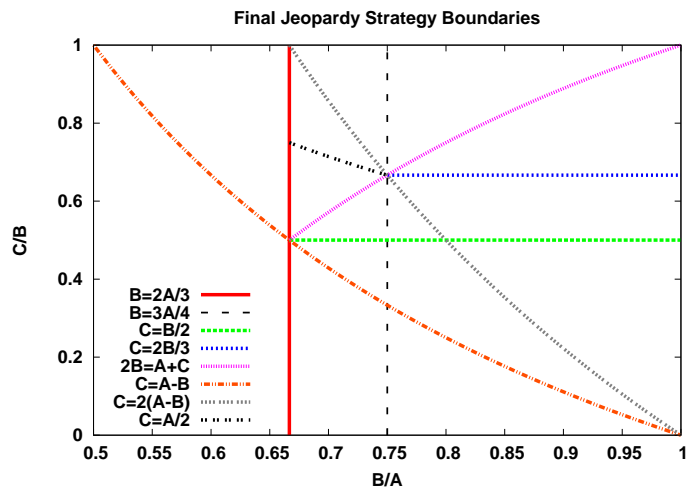

Figure 9: Illustration of important Final Jeopardy strategy regions and boundaries, as a function of B's score relative to A, and C's score relative to B.

## 4.2 More Aggressive DD Wagering

As we mentioned earlier in section 3.1.4, our analysis indicates that human contestants systematically err on the conservative side in DD wagering, given their actual likelihood of answering the DD clue correctly. This may reflect underestimation or ignorance of their likely DD accuracy, as well as a lack of quantitative means to estimate the impact of a score increase or decrease on one's overall winning chances. Another possibility is that contestants may recognize at some level that an aggressive bet is called for, but are too risk averse to actually try it.

In this section we present a specific historical example to illustrate how our analysis works, and to motivate the potential advantages of more aggressive wagering, as long as the player has reasonably good confidence in being able to answer correctly. Our example wager is taken from a J! Archive episode which aired during the last decade. The second place player "B" found the last DD on the $800 clue in a category where one previous clue ($400) had been played; this clue was answered correctly by B. At that point, the scores were all quite close, with B having 10400 while the opponents had 11400 and 9400. There were eight remaining clues to be played, worth a total of $10800. B chose to wager only $1000, got the DD right, and ultimately won the game.

We of course do not know B's confidence in this situation, but we suspect it was reasonably good, because: (a) $800 second-round DDs tend to be easy, with average contestant accuracy of about 72%; (b) B had already answered correctly the one previous clue in the category; (c) wagers of $1000 tend not to be indicative of unusually low DD accuracy. Given the above considerations, we suspect that B had at least a 70% chance of answering the DD correctly, and the low wager was due to a desire to avoid dropping into third place after an incorrect answer.

As one might surmise from reading section 3.1.4, our analysis suggests that at 70% confidence, the best wager is a True Daily Double, $10400. A plot of Monte-Carlo right/wrong equity curves, and equity at 70% confidence, as a function of amount wagered is shown in Figure 10. Note that for large bets, the red curve has a smaller magnitude slope than the green curve, and it decelerates more rapidly. Once the bet is sufficiently large, there is little incremental equity loss in increasing the bet, since the player has almost no chance to win at that point. Conversely, there is strong incremental gain from increasing the bet and getting the DD right. These factors tilt the calculation decidedly in favor of a maximal bet. Paradoxically, the almost certain loss after getting the DD wrong may be exactly why humans avoid betting True DDs in this situation. Many psychological studies have documented "irrational" preferences for taking immediate gains, or avoiding immediate losses, which have been attributed to so-called "hyperbolic discounting" (Ainslie, 2001). Since *Jeopardy!* is an undiscounted game, correcting any natural tendencies towards overly short-term thinking may be advisable for prospective contestants.

After a $1000 bet, B is either tied for first or tied for second, but the game is still very close so there is little change in equity. MC estimates B's equity at 39% if right and 31% if wrong. However, after a true DD wager, B would either obtain a commanding lead with 20800 and an estimated equity of 70% after answering correctly, or drop to zero with only about 3% equity after answering incorrectly. The equity difference between these two bets

is compelling at 70% confidence: betting $1000 gives 36.6% equity, while betting $10400 gives 49.9% equity, a hefty improvement of 13.3% in overall winning chances.

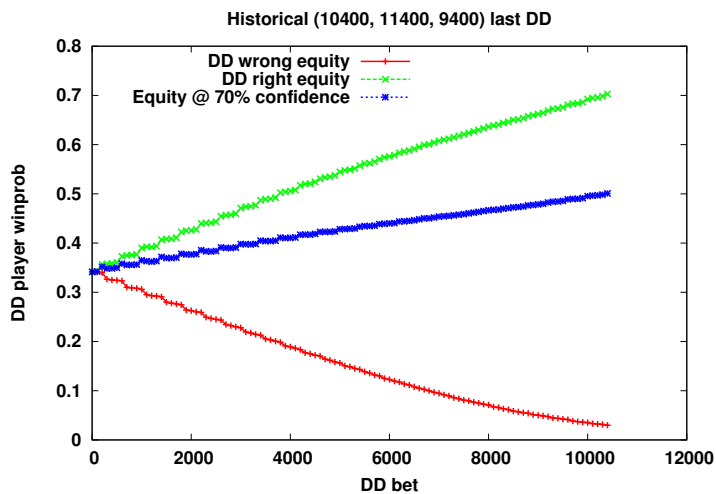

Figure 10: Equities after getting the DD right or wrong and at 70% confidence, in the example historical last-DD situation with scores (10400, 11400, 9400).

## 4.3 Counterintuitive Buzz-in Thresholds

Probably the most counterintuitive result of our analysis of buzz-in confidence thresholds is that attempting to answer may be correct even if a player has negative equity expectation in doing so. When we discussed WATSON's 50% default threshold with human contestants during the Sparring Games, many of them seemed surprised at such a low value, and some even objected vociferously. While their arguments were not based on quantitative equity estimates, they seem to intuitively recognize that WATSON's game standing would diminish on average after buzzing at 50% confidence, since WATSON's score change would be zero on average, but one of the opponent scores would likely increase. We tried to explain that this is clearly better than the alternative of not buzzing, where WATSON would again have zero expected score change, but the unimpeded opponents would have greater chances for a score increase.

Having developed an offline Monte-Carlo method for computing buzz-in thresholds for human contestants (see Appendix 2 for details), we will compare the actual MC results with a simple approximate formula for the threshold, derived below, which gives insight into how a negative-expectation threshold comes about. A more complex analytic calculation, yielding closed-form analytic expressions for all four threshold values, is detailed in the "Max-Delta approximation" section of Appendix 2. This analysis also agrees closely with our MC results.

We consider equities for one player (the "strategic" player) relative to a baseline state, with equity $E_{000}$, where the clue expires with no score change. We aim to calculate a confidence threshold $\theta_0$ on the initial buzz such that if the player's confidence $c = \theta_0$, then the equities buzzing and not buzzing are equal, i.e., $E_{NB}(c) = E_B(c)$.

Let $N = E_{NB} - E_{000}$ denote the expected equity change if the player does not buzz, either initially on any rebound. This will depend on how often the opponents buzz and their precision, but for good opponents, $N$ should be some negative value. In our MC simulations of early-game states with the refined Average Contestant model (where $b$ and $p$ are estimated based on round and clue row number), $N$ is approximately -0.86% for a \$1000 clue in the first round. In order to understand the effects of correlated opponent buzzing and precision, we also ran simulations with a corresponding uncorrelated model, obtaining $N \sim$-0.96% in the same situation.

Now consider the case where the player buzzes initially. For confidence values $c$ in the vicinity of $\theta_0$, if the player loses the buzz, we argue that the outcome should again be $N$: since rebound thresholds are higher than $\theta_0$, as shown below, the player will not buzz on any rebounds. Hence buzzing should only differ from not buzzing when the player wins the buzz. After winning the buzz and answering correctly, the player's equity will increase by some positive equity gain $G$. For early \$1000 first-round clues, $G$ appears to be $\sim$+3.25%, regardless of whether the opponents are correlated or uncorrelated with the player. After an incorrect answer, due to approximate linearity of equity with respect to score changes early in the game, the player will have an equity loss $\sim -G$, plus a likely further loss when the opponents play the rebound. For uncorrelated opponents, the extra loss should be about $N$, while for correlated opponents, it should be some lesser but still negative value $N'$, due to the fact that rebound precision is less than initial buzz precision in the correlated model. Assuming $N' = N$ for simplicity, confidence values $c$ where buzzing is better than not buzzing are determined by the following inequality:

$$cG + (1-c)(N-G) \geq N \tag{6}$$

Rearranging terms, we obtain: $c \geq \theta_0 = G/(2G-N)$. Since $N$ is negative, the confidence threshold will be less than $G/2G$, i.e., less than 50%. For the above quoted values of $G$ and $N$, equation 6 yields a threshold value of $\theta_0 = 0.436$ in the uncorrelated model, and $\theta_0 = 0.442$ in the correlated model.

We now compare with the actual MC threshold calculation, seen in Table 6, for one player (the leader) in a typical early game situation, where the first column has been played, the scores are (1800, -600, 1000), and the DD has not been played. Similar threshold values are robustly obtained in most early states, regardless of whether the player is leading or trailing. As per section 3.4, $\theta_0$ denotes the initial threshold, $\theta_1$ and $\theta_2$ denote the first rebound thresholds, and $\theta_3$ denotes the second rebound threshold. Note that the $\theta_0$ values for \$1000 clues in the correlated and uncorrelated models match very well with the approximate formula values. As they are also statistically equal, this suggests that the MC calculation for $\theta_0$ is robust in that there is no sensitive dependence on an assumed modest level of contestant correlation.

We also note an increase in first rebound thresholds for both models, and a further increase in second rebound thresholds. This makes sense as the expected loss when not buzzing, $N$, should diminish when the opponents are not eligible to buzz. For double rebounds, $N$ should equal 0, leading to a threshold of 0.5 according to the approximate formula. The increase in rebound thresholds is modest for the uncorrelated model, but quite significant for the correlated model. This is due to positive correlation of precision, implying that a player's posterior confidence is reduced after observing one or both opponents answer incorrectly.

Similar experiments for $200 clues obtain a more aggressive initial threshold (42% vs 44%). This is as expected: since $200 clues are easier, the opponents are more likely to buzz and answer correctly if the strategic player does not buzz. Hence the magnitude of $N$ relative to $G$ should increase, yielding a lower threshold. While not shown in Table 6, thresholds for $400, $600, and $800 clues take on plausible intermediate values between the $200 and $1000 limiting cases.

| clue value | $(\rho_b, \rho_p)$ | $\theta_0$ | $\theta_1$ | $\theta_2$ | $\theta_3$ |
|---|---|---|---|---|---|
| $1000 | (0.2, 0.2) | 0.44 | 0.67 | 0.68 | 0.78 |
| $1000 | (0, 0) | 0.44 | 0.49 | 0.50 | 0.53 |
| $200 | (0.2, 0.2) | 0.42 | 0.69 | 0.68 | 0.83 |
| $200 | (0, 0) | 0.42 | 0.47 | 0.48 | 0.54 |

Table 6: Early buzz-in thresholds in correlated and uncorrelated refined Average Contestant models, based on 800K MC trials of the 20 end-of-clue states. Test position scores (1800, -600, 1000), one column played, DD remains to be played.

In summary, while human contestants do not make precise confidence estimates, we suspect that their buzz attempts are safely above 50%, where they are more likely to be right than wrong. We would also be surprised if they became more cautious on rebounds, after one or both opponents answered incorrectly. By contrast, our analysis suggests that early in the game, it may be profitable to make slightly more speculative buzz attempts on the initial buzz, where the odds are even or slightly against getting it right. An important caveat is that such speculative guesses should not have a strong "tip-off" effect that would significantly aid a rebounder.

We would also advocate exercising caution on rebounds. Despite the tip-off effect, there is clear historical evidence that human precision is positively correlated, and declines on rebounds. As seen in our correlated threshold calculation, contestants should have well above 50% initial confidence to venture a rebound attempt, especially on a double rebound.

### 4.4 Lock-Tie Implications

We conclude this section by examining some of the strange and amusing consequences of the "Lock-Tie" scenario in Final Jeopardy, where B's score is exactly half of A's score. In this scenario, A is likely to bet nothing, so that B can achieve a tie for first place by betting everything and getting FJ right. This is decidedly preferable to having more than half of A's score, where B would need to get FJ right and A to get FJ wrong in order to win. The preference of B for a lower score can lead to some unusual strategy decisions (to say the least) near the end of the game. For example, Dupee (1998) discusses DD wagering on the last clue before Final Jeopardy, where the DD player has 7100 and the opponents have 9000 and 1000. Dupee advocates wagering $2600, which takes the lead with 9700 if correct, and drops to a lock-tie at 4500 if incorrect.

Watson's analysis turns up many such last-clue DD situations where lock-tie considerations lead to unusual or even paradoxical bets. For example, in episode 5516, Greg Lindsay[7] faced a last-clue DD decision, trailing with 6200 vs. 19200 and 9500. Greg wagered $6000 and ultimately won the game. However, Watson regards this as a serious error, and recommends wagering $3400 instead, which achieves a lock-tie at 9600 after a correct answer. We also frequently find Watson wagering more than necessary on a last-clue DD to achieve a lockout. In one such example, Watson had 26000 and the opponents had 19800 and 4400. Watson only needed to bet $13600 to secure the lockout, but this puts Watson at 12400 after answering incorrectly. Watson instead bet $16100, which also achieves a lockout if correct, but drops to a second-place lock-tie score of 9900 after answering incorrectly.

Watson also discovered that the lock-tie can influence wagering several squares before the end of the game. In our analysis of historical last-DD human bets, we found a class of situations where the DD player is trailing badly, and Watson recommends betting exactly $100 less than a True Daily Double. An example situation is where the DD player has 5000, the opponents have 21400 and 2800, and there are five remaining clues after the DD to be played, worth a total of $6000. Watson recommends betting $4900, which certainly seems weird and inconsequential, but there appears to be a real point to it. Note that the leader's score of 21400 happens to be an odd multiple of 200 (107x200). Since all remaining clues are even multiples of 200, the leader's score entering FJ will always be an odd multiple of 200. Now, in order to reach a lock-tie FJ, it follows that the DD player's score must be an odd multiple of 100. This is achieved by wagering $4900, with appreciable chances of a lock-tie, instead of $5000 which makes the lock-tie impossible. As the $4900 bet offers 7.2% equity instead of 6.4%, the lock-tie potential constitutes a substantial portion of overall winning chances in this situation.

Finally, we note that lock-ties can provide an incentive for players to intentionally give wrong answers. We were first alerted to this possibility by puzzle editor Peter Gordon, who emailed a last-clue DD scenario where Watson has 6000 and the opponents have 10000 and -1000. Peter recommended that Watson bet $1000 and get it wrong on purpose! Such a course of action would only require Watson to get FJ right in order to win, whereas after a large DD bet to take the lead, Watson needs to get both the DD clue and FJ right in order to win.

In subsequent testing of Watson's buzz-in strategy, we found a number of fascinating last-clue scenarios where Watson reported that buzz/wrong on the rebound offers better equity than either buzz/right or not buzzing. It turns out that these scenarios all occur when A=2B-V, where V is the value of the last clue, and C is out of contention. (As an example, suppose A=18800, B=10000, C=7600 and V=1200.) This situation allows B a double chance to achieve a lock-tie! First of all, B should never buzz initially, both because a wrong answer results in getting locked out, and because A may buzz and get it right, which results in a lock-tie. Additionally, A may buzz and get it wrong, reducing A's score to A-V = 2(B-V). If this happens, B can again reach a lock-tie by buzzing and answering incorrectly. This scenario is not as remote as one might think – it seems to occur about once per season – and in the majority of cases, the lock-tie is spoiled by incorrect behavior of B.

---

7. Greg Lindsay was the only contestant to win three Sparring Games against Watson.

## 5. Conclusions

By combining an original simulation model of *Jeopardy!* with state-of-the-art statistical learning and optimization techniques, we created a set of real-time game strategy algorithms that made Watson a much more formidable *Jeopardy!* contestant. As we have documented in detail, our strategy methods resulted in a significant boost in Watson's expected win rate in the Sparring Games and in the Exhibition Match, when compared with simple heuristic strategies. For DD wagering, our neural net method obtained a 6% improvement in win rate compared with our prior heuristic, and we estimate a further 0.6% improvement by using live Monte-Carlo analysis for endgame DDs. For FJ betting, simulations show a 3% improvement over a simple heuristic that always bets to cover when leading, and bets everything when trailing. As seen in Table 5, our best square selection method improves over heuristics by ∼3-9%, depending on how much DD seeking is done by the heuristic. We have no data on heuristic endgame buzzing, but a conservative guess is that our Approximate DP method would achieve ∼0.5% to 1% greater win rate. The aggregate benefit of these individual strategy improvements appears to be additive, since simulations put Watson's win rate at 50% using all baseline strategies, versus 70% using all advanced strategies.

There is also ample evidence that each of our strategy algorithms exceeds human capabilities in real-time decision making. Historical replacement shows that Watson's Best-Response FJ wagering clearly outperforms human wagers. Watson is also better at finding DDs than humans, as seen from the excess fraction of DDs found relative to its average board control. According to simulations, this translates into a greater overall win rate.

In the cases of Daily Double wagering and endgame buzzing, it is clear that humans are incapable in real time of anything like the precise equity estimates, confidence estimates, and complex calculations performed by our algorithms to evaluate possible decisions. Watson's error rate improves over humans by an order of magnitude in last-DD situations, as we saw in section 3.1.4. There is likely to be a lesser but still significant improvement on earlier DDs. It is difficult to identify human buzzing errors: a failure to buzz is indistinguishable from losing the buzz, and even if a contestant buzzes and is wrong, the decision may be correct at high enough confidence. We surmise that in most cases, human buzz errors are small. Our algorithm's main advantage is likely in handling special-case endgame states where humans can make major errors, such as mishandling a lock-tie or needlessly locking themselves out.

In addition to boosting Watson's results, our work provides the first-ever means of quantitative analysis applicable to any *Jeopardy!* game state and covering all aspects of game strategy. Consequently, we have unearthed a wealth of new insights regarding what constitutes effective strategy, and how much of a difference strategy can make in a contestant's overall ability to win. While we have illustrated numerous examples of such insights throughout the paper, we expect that even greater understanding of proper *Jeopardy!* strategy will be obtained by further development and deployment of algorithms based on our approaches. Just as top humans in classic board games (Chess, Backgammon, etc.) now use computer software as invaluable study aids, we envision that studying with *Jeopardy!* strategy software could become a vital part of contestant preparation to appear on the show. Toward that end, we are currently developing a version of our DD wager calculator to be deployed on J! Archive. This will nicely complement the existing FJ wager calculator, and will make our DD analysis widely accessible for study by prospective contestants.

For simulation modelers, perhaps the most important take-home lesson from our work on Watson is a reminder of the merits of starting from an approach based on extreme simplification. It is generally appreciated that simulation predictions may be insensitive to many low-level details. However, given the central role of the natural language clues and category titles in *Jeopardy!* gameplay, it is at least mildly surprising that successful simulation models may completely ignore the natural language content. One might have also thought that simple mean-rate models would be inadequate, as they fail to capture potentially important hot and cold streaks in specific categories, as well as variance across contestants in general QA ability. Such factors are apparently not critically important to model for purposes of optimizing Watson's strategies. Finally, it was not clear that we could adequately predict expected outcomes of Watson vs. two humans from scant live-game data. We had only crude estimates of relative buzzability, etc., and made no attempt to model the impact of Watson's unusual gameplay on human performance, or the tip-off benefit to humans when Watson answers incorrectly. Despite these limitations, the validation studies of section 2.7 demonstrate remarkably accurate predictions of performance metrics.

Looking beyond the immediate *Jeopardy!* domain, we also foresee more general applicability of our high-level approach to coupling Decision Analytics to QA Analytics, which consists of building a simulation model of a domain (including other agents in the domain), simulating short-term and long-term risks and rewards of QA-based decisions, and then applying learning, optimization and Risk Analytics techniques to develop effective decision policies. We are currently investigating applications of this high-level approach in health care, dynamic pricing, and security (i.e., counter-terrorism) domains.

## Acknowledgments

We are grateful to the entire worldwide team at IBM that made the Watson project possible, the team at Sony/JPI that made the Exhibition Match and Sparring Games possible, and all the former *Jeopardy!* contestants who volunteered to participate in live test matches with Watson. We thank the anonymous reviewers and Ed Toutant for many helpful comments and suggestions to improve the manuscript.

## Appendix A. Watson's Competitive Record

Prior to appearing on *Jeopardy!*, Watson played more than 100 "Sparring Games" against former *Jeopardy!* contestants in a realistic replica of a TV studio, which was constructed at the IBM Research Center in Yorktown Heights, NY. The studio featured real *Jeopardy!* contestant lecterns and signaling devices, and made use of the actual JPI (Jeopardy Productions Inc.) game-control system. The content for each game (categories, clues, answers) was supplied directly by JPI, and consisted of actual *Jeopardy!* episodes that had already been taped, but not yet aired. (This eliminated the possibility that contestants could have previously seen the content used in their games.) A professional actor, Todd Crain, was hired to host the games. To incentivize the contestants, they were paid $1000 for each first-place finish, and $250 for each second-place finish.

An initial series of 73 games took place between Oct. 2009 and Mar. 2010. Contestants were recruited by JPI, most of whom had appeared on the show only once or twice, and none had appeared in more than three episodes. We considered these players as representative of "average" human contestants that appear on the show. Results in this series of games were that WATSON finished first in 47 games (64.4%), second in 15 games (20.5%), and third in 11 games (15.1%). We also note that 21 of WATSON's 47 wins were by "lockout," i.e., guaranteed wins where WATSON could not be caught in Final Jeopardy.

WATSON additionally played a second series of 55 games during Fall 2010, this time against much stronger human opposition. These were contestants who had competed in the show's annual Tournament of Champions, and had done well enough to reach the final or semi-final rounds. WATSON was also considerably improved in all respects, and in particular, its full complement of advanced quantitative strategies was deployed in these games. (By contrast, the only advanced strategy in most of the Series 1 games was for Final Jeopardy betting.) Results in this series were as follows: WATSON finished first in 39 games (70.9%), second in 8 games (14.55%) and third in 8 games (14.55%). WATSON's rate of winning by lockout also improved, to 30 out of 39 games (76.9%), vs. $21/47 = 44.7\%$ in the previous series.

Finally, as witnessed by millions of viewers, WATSON played a two-game Exhibition match against Ken Jennings and Brad Rutter, arguably the two best human *Jeopardy!* contestants of all time. WATSON took the $1,000,000 first-place prize by lockout, with a total score of 77,147. Ken Jennings took second place ($300,000) with a score of 24,000, and Brad Rutter finished in third place ($200,000) with a score of 21,600.

## Appendix B. Buzz Threshold Calculation Details

This Appendix presents computational details of our method for calculating initial buzz and rebound buzz decisions in an endgame state with no remaining Daily Doubles, a current selected square with dollar value $D$, and $K$ remaining squares to be played after the current square. We assume that we are optimizing the buzz decision of one player (the "strategic" player), and that the two opponents are "non-strategic" players in that their buzz decisions are determined by some fixed stochastic process. We further assume that the opponents' buzz decisions do not change going from the initial buzz to a rebound or second rebound. By default the strategic player is WATSON, although we have also developed a similar method to compute confidence thresholds for human contestants.

The calculation as invoked by WATSON in live play assumes that WATSON's confidence is not yet known, since computation must begin before the QA system has returned a confidence value. We therefore pre-compute buzz decisions over discretized confidence values between 0 and 1 with discretization interval typically set to 0.01.

### B.1 Calculation for WATSON vs. Two Humans

Calculation proceeds by diagramming a tree of possible events starting from the initial buzz state and leading to all possible end-of-clue states, each corresponding to a different score change combination. We denote the end-of-clue equities by $E_{xyz}$, where the first index denotes Watson's score change, and possible values of $x$, $y$ and $z$ are "+" (the player's score increased by $D$), "0" (score remained unchanged), and "-" (score decreased by $D$). Since at most one contestant can have a score increase, there are a total of 20 end-of-clue states: 12 where a contestant got the clue right, and eight where no one got it right.

The tree allows for the possibility that Watson may buzz or may not buzz in each of the four live states (0=initial buzz, 1=rebound with human #1 wrong, 2=rebound with human #2 wrong, 3=rebound with both humans wrong) where Watson is eligible to buzz. Descending the tree starting from the initial buzz state, it assigns transition probabilities to every branch, using our regular-clue model of human performance, along with probabilities that Watson will win a contested buzz when one or two humans are attempting to buzz.

Having defined the tree, the transition probabilities, and a set of end-of-clue states, the algorithm first estimates the end-of-clue equities by Monte-Carlo trials over the remaining clues and Final Jeopardy. The MC trials make use of the stochastic process models of human and Watson performance on regular clues that we presented earlier.

One useful trick we employ here is to reuse each MC trial over remaining clues in each of the 20 end-of-clue states, instead of generating independent trials in those states. This is done by first performing a trial in the $(0, 0, 0)$ state, where no player attempted to buzz in, and then offsetting the sequence of scores and the scores going into FJ by the specific score change of each end-of-clue state. This enables faster computation and achieves more statistically significant comparisons between states than would result from independent trials. Additionally, while each trial is being performed, we monitor at each step whether Watson has achieved a guaranteed lockout given the starting scores and the specific score change combinations of each end-of-clue state. If so, we mark the trial as a guaranteed win for that end-of-clue state: this obviates the need to simulate FJ in that trial, and makes the simulations more faithful, since Watson actually uses lockout-preserving buzzing in live play.

Having evaluated the end-of-clue states as described above, the calculation works backwards to evaluate progressively higher interior tree nodes. We first calculate confidence-independent values of the Watson-ineligible states where Watson buzzes and is wrong. There are three such states: IS0 (Watson wrong on the initial buzz), IS1 (Watson wrong after human #1 is wrong) and IS2 (Watson wrong after human #2 is wrong). Formulas for these values are written below.

To establish notation in these formulas, recall that our human models generate correlated binary events at the start of a regular clue indicating whether the contestants attempt to buzz, and whether they have a correct answer. These binary variables persist as the clue is played, so that their buzz decisions and correctness do not change during rebounds. With this in mind, we let $b_{00}, b_{01}, b_{10}, b_{11}$ denote the probabilities of the four possible buzz/no-buzz joint decisions for a pair of humans, and $p_{00}, p_{01}, p_{10}, p_{11}$ denote probabilities of the four possible joint right/wrong outcomes. We typically assume symmetric human models where $b_{01} = b_{10}$ and $p_{01} = p_{10}$. We further let $p_H = p_{10} + p_{11} = p_{01} + p_{11}$ denote the single-contestant human precision, and $b_H = b_{10} + b_{11} = b_{01} + b_{11}$ denote the single-contestant human buzz attempt rate. These values may be fixed for all clue values, or we may use estimated values that depend on round and row number, as depicted in Figure 5.

With the above notation, the formula for IS0 value is:

$$
\begin{aligned}
V(IS0) \;=\; & b_{00}E_{-00} + p_H(b_{10} + b_{11}/2)(E_{-+0} + E_{-0+}) + (1 - p_H)b_{10}(E_{--0} + E_{-0-}) \\
& + p_{01}b_{11}(E_{--+} + E_{-+-})/2 + p_{00}b_{11}E_{---}
\end{aligned}
\tag{7}
$$

Similarly, values of the IS1 and IS2 states are given by:

$$
V(IS1) = \frac{b_{10}}{b_H}E_{--0} + \frac{b_{11}}{b_H}\left(\frac{p_{01}E_{--+} + p_{00}E_{---}}{1 - p_H}\right)
\tag{8}
$$

$$
V(IS2) = \frac{b_{01}}{b_H}E_{-0-} + \frac{b_{11}}{b_H}\left(\frac{p_{10}E_{-+-} + p_{00}E_{---}}{1 - p_H}\right)
\tag{9}
$$

Note in the above expressions that we require conditional probabilities for the remaining eligible human to buzz and answer correctly, given that the first human buzzed and answered incorrectly. Using unconditional probabilities would correspond to a model that re-draws buzz/no-buzz and right/wrong outcomes on each rebound, which is not consistent with our model.

Next we evaluate the live states LS0, LS1, LS2, LS3 where WATSON is eligible to buzz, starting from the double-rebound state LS3, and working backwards to the first rebound states LS1 and LS2, and finally the initial-buzz state LS0. We compute separate evaluations in the cases where WATSON buzzes or does not buzz; the larger of these determines the optimal policy and optimal value function.

At a given WATSON confidence level $c$, the values of the double-rebound state when WATSON buzzes or does not buzz are given respectively by:

$$
V_B(LS3, c) \;=\; cE_{+--} + (1 - c)E_{---}
\tag{10}
$$

$$
V_{NB}(LS3, c) \;=\; E_{0--}
\tag{11}
$$

Thus WATSON's optimal buzz decision $B^*(LS3, c) = \arg\max_{B,NB}\{V_B(LS3, c), V_{NB}(LS3, c)\}$ and the optimal state value is: $V^*(LS3, c) = \max\{V_B(LS3, c), V_{NB}(LS3, c)\}$.

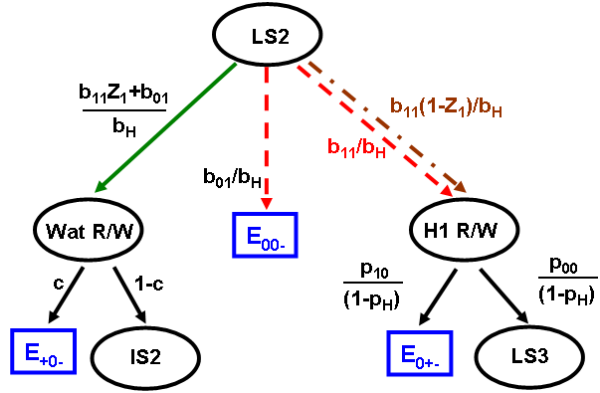

Figure 11: Event tree for live state 2, where human #2 buzzed initially and was wrong.

Calculation of the first rebound state values proceeds as diagrammed in Figure 11, which considers the LS2 state where human #2 buzzed first and answered incorrectly. Red arrows indicate outcomes where WATSON does not buzz. Green and brown arrows indicate respective cases where WATSON wins the buzz and loses the buzz. $Z_1$ denotes the probability that WATSON wins a contested buzz against one human. The analogous tree for LS1 is obtained by interchanging human #1 ↔ #2 indices.

$$V_{NB}(LS2, c) = \frac{b_{01}}{b_H}E_{00-} + \frac{b_{11}}{b_H}\left[\frac{p_{10}E_{0+-} + p_{00}V^*(LS3, c)}{1 - p_H}\right] \tag{12}$$

$$V_B(LS2, c) = \frac{b_{11}Z_1 + b_{01}}{b_H}\left[cE_{+0-} + (1-c)V(IS2)\right]$$

$$+ \frac{b_{11}(1 - Z_1)}{b_H}\left[\frac{p_{10}E_{0+-} + p_{00}V^*(LS3, c)}{1 - p_H}\right] \tag{13}$$

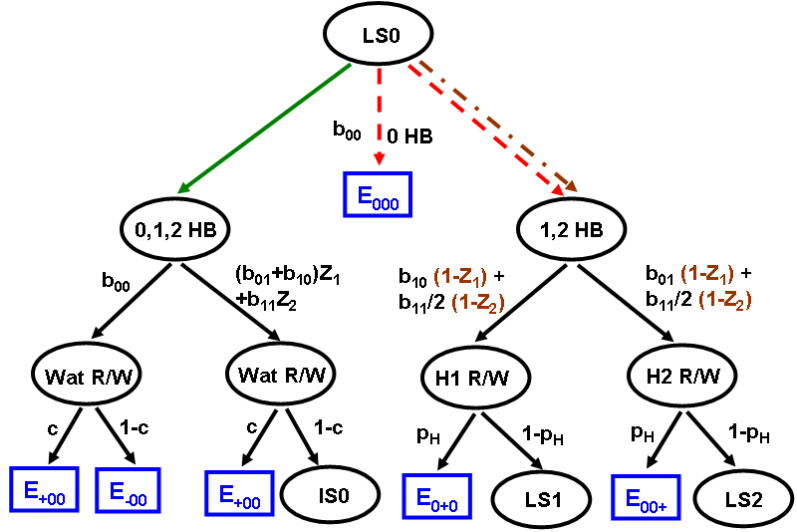

Figure 12: Event tree for live state 0, i.e., the initial buzz state.

Finally, Figure 12 illustrates the analysis of the initial buzz state LS0. $Z_2$ denotes the probability that WATSON wins the buzz when both humans are buzzing.

$$V_{NB}(LS0, c) = b_{00}E_{000} + (b_{10} + b_{11}/2)[p_H(E_{0+0} + E_{00+}) +$$

$$(1 - p_H)(V^*(LS1, c) + V^*(LS2, c))] \tag{14}$$

$$V_B(LS0, c) = b_{00}(cE_{+00} + (1-c)E_{-00})$$

$$+ ((b_{01} + b_{10})Z_1 + b_{11}Z_2)(cE_{+00} + (1-c)V(IS0))$$

$$+ (b_{10}(1 - Z_1) + b_{11}(1 - Z_2)/2)(p_HE_{0+0} + (1 - p_H)V^*(LS1, c))$$

$$+ (b_{01}(1 - Z_1) + b_{11}(1 - Z_2)/2)(p_HE_{00+} + (1 - p_H)V^*(LS2, c)) \tag{15}$$

## B.2 Calculation for Human vs. Two Humans

Here we present an extension of the above calculation where the "strategic" player is a human instead of WATSON. This scenario introduces additional complexity in that, unlike WATSON, the strategic player's performance is correlated with that of the opponents (and vice versa).

Our approach hypothesizes a mechanism to generate correlated private confidence estimates $(c_0, c_1, c_2)$ for each player when the current clue is revealed, drawn from a suitable multi-variate confidence distribution. We assume that the non-strategic players attempt to buzz in when their private confidence value exceeds a fixed threshold, chosen so that the probability mass above the threshold matches the desired attempt rate, and the first moment above the threshold matches the desired precision. In the uniform $(b, p)$ model over all clues, as described in section 2.4, we would match the target values $b = 0.61, p = 0.87$ by using Beta distribution for each player, $Beta(0.69, 0.40)$, with a buzz threshold value set to 0.572. However, we obtain a more accurate and meaningful threshold model for humans by fitting a different Beta distribution to each $(b, p)$ parameter combination estimated by round and row number, as plotted in Figure 5.

We obtain correlated draws from the resulting multi-variate Beta distribution via the "copula" technique (Nelsen, 1999). This entails drawing $\vec{x} = (x_0, x_1, x_2)$ from a suitably correlated multi-variate normal distribution, mapping to the respective CDF values $\vec{X} = (X_0, X_1, X_2)$ which lie in the unit interval, and then mapping these values to the inverse CDF of the Beta distribution. Since the confidence draws are correlated, this will result in correlated buzzing by the non-strategic players. We further obtain correlated precision by similarly generating correlated uniform numbers in the unit interval to compare with the players' confidence values as a basis for assigning right/wrong answers. A correlation coefficient of $\sim 0.4$ matches the observed precision correlation in Average Contestants.

With such a model of correlated confidence-based buzzing and precision, we are now equipped to make necessary modifications to the calculations described in Equations 7-15 and Figures 11,12. In every case, the opponent attempt rates and precisions need to be conditioned on the strategic player's confidence value $c$. We can make accurate numeric estimates of these conditional probabilities by running many millions of trials with the simulation model, discretizing the observed $c$ in each trial, and recording at each discrete level the number of times that 0, 1 or 2 opponents buzzed, and 0, 1, or 2 opponents had a correct answer. Additionally, when considering buzzing on a first or second rebound, the strategic player needs to estimate a posterior confidence given that one or two opponents have already buzzed and answered incorrectly. This can result in a significant drop in estimated confidence: for example, an initial confidence of 80% will drop to a posterior value of only 50% in the double-rebound state LS3. Finally, in any of the ineligible states IS0, IS1, IS2, the expected opponent precisions must also be conditioned upon the strategic player buzzing and answering incorrectly.

## B.3 Max-Delta Approximation

We developed a greatly simplified analytic approximation to the calculations given in Equations 7-15 by making the following assumptions: (i) the current game state is far from the end of the game (i.e., the current clue value $D$ is much smaller than the total value of all remaining clues); (ii) all three players have intermediate probabilities of winning (i.e., not close to 0 or 1). Under these assumptions, a player's equity change at the end of the clue should be approximately linear in the score changes of the three players. Intuitively, we may write: $\Delta E \sim (\chi_1 \Delta(S_0 - S_1) + \chi_2 \Delta(S_0 - S_2))$, where $S_0$ is the player's score and $S_1, S_2$ are the opponent scores, recognizing that the chances of winning depend on score positioning relative to the opponents. If we further assume that the opponent scores are similar (which is often true early in the game), we then have $\chi_1 \simeq \chi_2 \simeq \chi$, an overall scaling factor. As the clue value $D$ is also an overall scaling factor, we can express the Max-Delta objective of the buzz decision by rewriting the end-of-clue equities $E_{xyz}$ appearing in Equations 7-15 as $E_{xyz} = 2x - y - z$.

To further facilitate analytic calculation, we also assume that the opponents' buzz attempts and the precisions of the three players are all uncorrelated. The equities of the ineligible states {IS0,IS1,IS2} then reduce to:

$$
\begin{aligned}
V(IS0) &= -2(1-b)^2 - 6bp(1-b/2) - 2(1-p)b(1-b) - 4p(1-p)b^2 \\
V(IS1) &= V(IS2) = b - 1 - 2bp
\end{aligned}
\tag{16}
$$

We can similarly rewrite the equities in the four live states {LS0,LS1,LS2,LS3} after buzzing or not buzzing. For the double-rebound state LS3, these reduce to $V_B(LS3) = 4c$ and $V_{NB}(LS3) = 2$. At the threshold confidence $c = \theta_3$, we have $4\theta_3 = 2$, so that $\theta_3 = 0.5$. By likewise equating the buzz/no-buzz equities in the other live states, we can obtain closed-form analytic expressions for the respective thresholds $\{\theta_0, \theta_1, \theta_2\}$. For the first-rebound thresholds, we have:

$$
\theta_1 = \theta_2 = \frac{2 + b^2(1-Z_1)(1-2p) + b(2p-3)(1-Z_1)}{(b(Z_1-1)+1)(b(2p-1)+4)}
\tag{17}
$$

This expression assumes that $\theta_1, \theta_2 \leq \theta_3$ so that the player will not buzz in the second-rebound state. Finally, the initial buzz threshold $\theta_0$ (again assuming no-buzz in the rebound states) is:

$$
\theta_0 = \frac{wm + 2(1-b)^2 - 2wV(IS0)}{4(1-b)^2 + 4w - 2wV(IS0)}
\tag{18}
$$

where $m = -2p + 2(1-p)(1+b-2bp)$ is the total equity change after losing the buzz to either opponent, and $w = b(1-b)Z_1 + 0.5b^2 Z_2$ is the probability of beating one opponent on the buzzer.

For average human contestants, we set $b = 0.61$, $p = 0.87$, $Z_1 = 1/2$, and $Z_2 = 1/3$, yielding first-rebound thresholds $\theta_1 = \theta_2 = 0.478$ and initial buzz threshold $\theta_0 = 0.434$. The four Max-Delta thresholds computed here are quite close to the uncorrelated Monte-Carlo values reported in Table 6 for simulations of average contestants in early game states.